\pdfoutput=1
\documentclass[11pt]{article}

\usepackage[]{acl}

\usepackage{times}
\usepackage{latexsym}

\usepackage[T1]{fontenc}
\usepackage[utf8]{inputenc}

\usepackage{color}
\usepackage{tabularx}
\usepackage{tcolorbox}
\usepackage{xcolor}
\usepackage{colortbl}
\usepackage{tabu}
\usepackage{booktabs}
\usepackage{multirow}

\usepackage{graphicx} 
\usepackage{float}
\usepackage{subfigure} 
\usepackage{amssymb}

\usepackage{hyperref}
\usepackage{tablefootnote}
\usepackage{xspace}

\usepackage{float}
\usepackage{amsmath}
\usepackage{bm}
\usepackage{algorithmicx}
\usepackage{algorithm}
\usepackage{algpseudocode}

\usepackage{arydshln}

\usepackage{amssymb}
\usepackage{pifont}
\newcommand{\cmark}{\ding{51}}%
\newcommand{\xmark}{\ding{55}}%
\usepackage{enumitem}
\usepackage{fontawesome5}

\usepackage{listings}
\usepackage{microtype}

\newcommand{\ours}{Finance\textsc{Math}\xspace}
\newcommand{\nexample}{1,200\xspace}
\newcommand{\nknowledge}{864\xspace}
\newcommand{\nmodel}{51\xspace}
\newcommand{\norg}{16\xspace}

\newcommand{\dev}{\emph{development}\xspace}
\newcommand{\eg}{\hbox{\emph{e.g.,}}\xspace}
\newcommand{\ie}{\hbox{\emph{i.e.,}}\xspace}

\newcommand{\up}[1]{\textcolor{red}{(+#1)}}
\newcommand{\down}[1]{\textcolor{blue}{(-#1)}}

\title{\ours: Knowledge-Intensive Math Reasoning \\in Finance Domains}
\author{Yilun Zhao\thanks{~~Equal Contribution}~~$^{1}$ \quad Hongjun Liu$^{*2,3}$ \quad Yitao Long$^3$ \\ \bf{Rui Zhang$^4$ \quad Chen Zhao$^{2,3}$ \quad Arman Cohan$^{1,5}$} \vspace{4pt}\\
$^1$Yale University \quad $^2$NYU Shanghai \quad $^3$New York University \\ $^4$Penn State University \quad $^5$Allen Institute for AI\\\newline \vspace{10pt}
}

\begin{document}
\maketitle

\begin{minipage}[t]{2\linewidth}
\vspace{-1.75cm}
  \centering
  \href{https://github.com/yale-nlp/FinanceMath}{{\faGithub{}}\xspace\texttt{github.com/yale-nlp/FinanceMath}} \\\vspace{2pt}
  \href{https://financemath-acl2024.github.io}{{\faGlobe{}}\xspace\texttt{financemath-acl2024.github.io}} \\
\vspace{0.5cm}
\end{minipage}

\begin{abstract}
We introduce \ours, a novel benchmark designed to evaluate LLMs' capabilities in solving knowledge-intensive math reasoning problems. 
Compared to prior works, this study features three core advancements.
First, \ours includes \nexample problems with a hybrid of textual and tabular content. These problems require college-level knowledge in the finance domain for effective resolution. 
Second, we 
provide expert-annotated, detailed solution references in Python program format, ensuring a high-quality benchmark for LLM assessment. We also construct a finance-domain knowledge bank and investigate various knowledge integration strategies.
Finally, we 
evaluate a wide spectrum of \nmodel LLMs with both Chain-of-Thought and Program-of-Thought prompting methods. Our experimental results reveal that the current best-performing system (\ie GPT-4o) achieves only 60.9\% accuracy using CoT prompting, leaving substantial room for improvement. Moreover, while augmenting LLMs with external knowledge can improve model performance (\eg 47.5\% $\rightarrow$ 54.5\% for Gemini-1.5-Pro), their accuracy remains significantly lower than the estimated human expert performance of 92\%.
We believe that \ours can advance future research in the area of domain-specific knowledge retrieval and integration, particularly within the context of solving reasoning-intensive tasks.

\end{abstract}

\begin{figure}[!t]
    \centering
    \includegraphics[width = \linewidth]{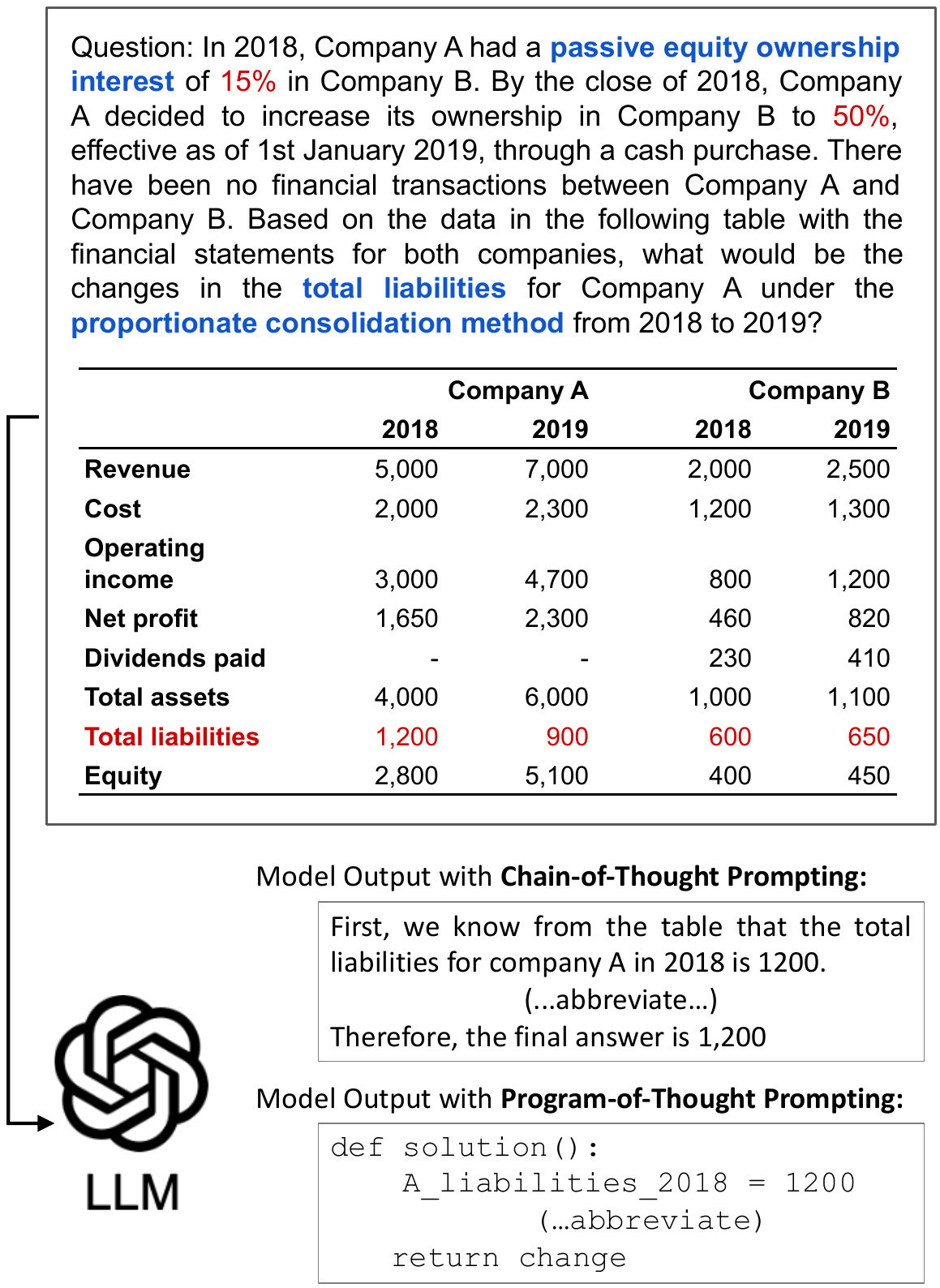}
    \caption{An example of \ours. To answer the given question, LLMs are required to comprehend specialized financial terms, such as ``passive equity ownership interest'' and ``proportionate consolidation method''. Additionally, they must interpret tabular data within the question and accurately identify question-relevant data points in the table.}
    \label{fig:example}
\end{figure}
\begin{table*}[!t]
\centering
\resizebox{\textwidth}{!}{%
\addtolength{\tabcolsep}{-0.1em}
\renewcommand{\arraystretch}{1.1}

\begin{tabular}{@{}llllrccl@{}}
\toprule
\multirow{2}{*}{Dataset}               & \multirow{2}{*}{Domain} & \multirow{2}{*}{Level} & \multirow{2}{*}{Source} & \multirow{2}{*}{\# Examples} & Table                          & Knowledge-                     & \multirow{2}{*}{Solution Format}       \\
                                       &                         &                        &                         &                               & Reasoning? & Intensive? & \\ \midrule
MAWPS~\cite{koncel-kedziorski-etal-2016-mawps}           & Math              & Elem. School      & Generated               &  3,320                               & \xmark                                & \xmark                              & Text                \\
ASDiv~\cite{miao-etal-2020-diverse}           & Math     & Elem. School      & Internet                & 2,305                               & \xmark                                & \xmark                              & Math Equation                \\
SVAMP~\cite{patel-etal-2021-nlp}              & Math     & Elem. School      & ASDiv                   & 1,000                              & \xmark                                & \xmark                              & Math Equation                \\
Math23K~\cite{wang-etal-2017-deep}            & Math                 & Elem. School      & Internet                & 23,162                              & \xmark                                & \xmark                              & Math Equation               \\
GSM8K~\cite{cobbe2021training}         & Math     & Middle School          & CrowdSource               & 8,500                              & \xmark                               & \xmark                             & Text               \\
MATH~\cite{hendrycks2021measuring}        & Math                    & High School            & Competition             & 12,500                              & \xmark                               & \xmark                             & Text               \\
AQuA~\cite{ling-etal-2017-program}            & Math          & College                & GMAT, GRE               & 100,000                              & \xmark                               & \xmark                             & Text               \\
MathQA~\cite{amini-etal-2019-mathqa}          & Math          & College                & AQuA                    & 100,000                              & \xmark                               & \xmark                             & Math Equation                 \\
MathQA-Python~\cite{austin2021program} & Math          & College                & AQuA                    & 23,914                              & \xmark                               & \xmark                             & Python Program               \\
MathVista~\cite{lu2023mathvista} & Math & Elem. to College & Internet+Expert & 6,141 & Few & Few & Text \\
\noalign{\vskip 0.5ex}\hdashline\noalign{\vskip 0.5ex}
TabMWP~\cite{lu2023dynamic}            & Math     & Middle School          & Textbooks               & 38,431                              & \cmark                            & \xmark                             & Text               \\
FinQA~\cite{chen-etal-2021-finqa}             & Finance                 & College                & Expert             & 8,281                              & \cmark                               & \xmark                             & Math Program               \\
TAT-QA~\cite{zhu-etal-2021-tat}               & Finance                 & College                & Expert             & 16, 552                              & \cmark                               & \xmark                             & Text               \\
MultiHiertt~\cite{zhao-etal-2022-multihiertt}                            & Finance                 & College                & Expert             &  10,440                              & \cmark                               & \xmark                          & Math Equation               \\
DocMath-Eval~\cite{zhao2023docmatheval} & Finance & College & Expert & 5,974 & \cmark                               & Few & Python Program \\
\noalign{\vskip 0.5ex}\hdashline\noalign{\vskip 0.5ex}
TheoremQA~\cite{chen-etal-2023-theoremqa}                              & STEM                    & College                & Internet+Expert         & 800                              & \xmark                               & \cmark                            & Text           \\ \midrule
% MMMU~\cite{yue2023mmmu} & STEM & College & \\
\ours (ours)                                 & Finance                 & College                & Internet+Expert         & \nexample                              & \cmark                               & \cmark                & Python Program

\\\bottomrule
\end{tabular}
}
\caption{Comparison between \ours and existing \textbf{math reasoning} datasets. \ours is distinguished by three unique characteristics: (1) \emph{Knowledge-Intensive}: Problems necessitate domain-specific knowledge, complemented by a financial knowledge bank for research facilitation; (2) \emph{Table Reasoning}: 40.2\% of problems incorporate table information, requiring models to understand table structure as well as interpret and reason over tabular data; (3) \emph{Expert Annotation}: Each problem is accompanied by a detailed, expert-annotated Python-formatted solution. Such solution annotation combines the explicitness of code execution with the descriptive power of natural language explanations in python comment format, offering a more effective and adaptable solution representation for complex math reasoning problems in \ours.}
\label{tab:dataset_comparison}
\end{table*}
\section{Introduction}
Large language models (LLMs) have been increasingly recognized for their potential for complex problem-solving in real-world scenarios~\cite{OpenAI2023GPT4TR, Touvron2023Llama2O, jiang2023mistral}. 
Solving math reasoning problems has emerged as a key method for assessing LLMs' capabilities~\cite{roy-roth-2015-solving, amini-etal-2019-mathqa, cobbe2021training, chen-etal-2023-theoremqa}, as it demands both understanding contextual information and reasoning over complex logics.

Recent advancements in LLMs have led to remarkable progress in solving fundamental math problems~\cite{wei2022chain, lewkowycz2022solving, chen2023program, wang2023selfconsistency, luo2023wizardmath, azerbayev2023llemma}. 
However, as illustrated in Table~\ref{tab:dataset_comparison}, existing math reasoning benchmarks 
typically do not require specialized domain knowledge. 
This becomes a notable shortcoming when considering practical applications of LLMs. 
Measuring progress in specialized areas such as finance and healthcare typically involves addressing \emph{domain-specific} and \emph{knowledge-intensive} problems, which goes beyond the scope of general mathematical reasoning.
Recognizing this gap in the existing benchmarks, we focus on the finance domain. We chose this domain because, as illustrated in \autoref{fig:example}, it often involves scenarios requiring not only basic mathematical skills but also a deep understanding of financial concepts~\cite{yang2023fingpt, xie2023pixiu, Wu2023BloombergGPTAL}. 
Additionally, the finance domain frequently employs tables to represent data~\cite{zhu-etal-2021-tat, chen-etal-2021-finqa, zhao-etal-2022-multihiertt, li-etal-2022-learning, li-etal-2022-finmath, zhao-etal-2023-qtsumm, zhao-etal-2023-robut}, which adds another layer of complexity to the knowledge-intensive problem-solving.

We introduce \ours, the first benchmark tailored for evaluating LLMs in the context of knowledge-intensive math reasoning in the Finance domain.
The dataset contains \nexample problems that cover a broad range of finance subareas, with 40.2\% of the problems necessitating data interpretation over tabular data. Each problem is accompanied by expert-annotated, Python-formatted solutions, providing a comprehensive reference for evaluating the LLMs' performance. Additionally, we collect and release a comprehensive knowledge bank, which includes detailed definitions and explanations for \nknowledge financial terms and concepts, facilitating future research on improving knowledge-intensive problem-solving through knowledge retrieval.

We evaluate a wide spectrum of open-source and proprietary LLMs, specifically, \nmodel model models from \norg organizations. Notably, this includes \emph{math-specific}~\cite{luo2023wizardmath, shao2024deepseekmath, ying2024internlmmath}, \emph{code-based}~\cite{guo2024deepseekcoder, luo2023wizardcoder, codestral, lozhkov2024starcoder2} LLMs, as well as \emph{mixture of experts} (MoE) LLMs~\cite{mistral2023moe, dbrx}.
Two prompting methods, Chain-of-Thought~\cite{wei2022chain} and Program-of-Thought~\cite{chen2023program}, are adopted for experiments. 

Our experimental results demonstrate a significant gap between existing LLMs and human experts. Specifically, the current best-performing system (\ie GPT-4o) achieves only 60.9\% accuracy with CoT prompting, which still lags far behind human expert performance in the open-book setting, which stands at 92\%. These results highlight the challenges of \ours, underscoring the need for further advancements in LLMs for knowledge-intensive problem-solving capabilities.
Next, we investigate how to integrate domain-specific knowledge to enhance the problem-solving capabilities of LLMs. We investigate various popular knowledge integration strategies and reveal that including question-relevant knowledge into the prompt can consistently improve LLMs' performance. This provides insights for future work to develop more advanced knowledge-augmented strategies to realize higher performance gains.

Our contributions are summarized below:
\begin{itemize} [leftmargin=*]
\itemsep0em 
\item We propose \ours, the first knowledge-intensive math reasoning benchmark in finance domains, aimed at evaluating LLMs' abilities in knowledge-intensive math reasoning.
\item We conduct comprehensive evaluations using a diverse array of LLMs, uncovering a substantial performance gap between the best-performing LLM (\ie GPT-4o) and human experts.
\item We present a detailed analysis on augmenting LLMs with various knowledge integration strategies. This provides valuable insights for future work in knowledge-intensive problem solving.
\end{itemize}
\section{\ours Benchmark}
In this section, we describe the dataset construction process for \ours. We begin by constructing a knowledge bank that includes well-formulated definitions of \nknowledge financial terms. We then instruct expert annotators to use knowledge terms within the constructed knowledge bank to create knowledge-intensive questions with a hybrid of textual and tabular content. 

\subsection{Knowledge Bank Construction}
We construct a knowledge bank that covers a wide range of \nknowledge knowledge terms in the finance domain. It simplifies the creation of knowledge-intensive questions by annotators and enables the exploration of various topics within domain knowledge.
The knowledge bank includes finance-domain-specific terms (\eg ``exchange rate'' and ``net present value'') collected from Wikipedia. Each knowledge term is accompanied with their corresponding \emph{textual definitions} and, where applicable, \emph{mathematical formulas} in python format. An example of included knowledge terms is illustrated in \autoref{fig:knowledge_example}. 
We detail the the main processes for knowledge bank construction as follows:

\begin{figure}[!t]
    \centering
    \includegraphics[width = \linewidth]{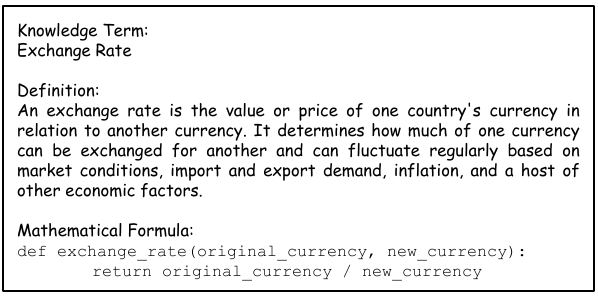}
    \caption{An example of knowledge terms ``Exchange Rate” included in the constructed knowledge bank.}
    \label{fig:knowledge_example}
\end{figure}
\paragraph{Knowledge Collection}
To construct a knowledge bank, we first collect knowledge relevant to the finance domain from Wikipedia 
using ``finance'' and ``economics'' as key search terms. After collecting the raw financial data, we adopt comprehensive heuristics, embedding-based methods to remove duplicates. This procedure ensures the uniqueness of each knowledge term in our bank. 

\paragraph{Automatic Knowledge Formulation}
To enhance the adaptability and usability of the knowledge bank, we incorporate a two-step automatic knowledge formulation process, making each piece of collected knowledge standardized and distilled into a clear, concise format. The primary motivation for using \emph{automatic} knowledge formulation is cost efficiency and effectiveness. We have observed that GPT-3.5 models are adept at handling this straightforward task with minimal bias, as this process does not involve the addition of extraneous knowledge.
We first prompt GPT-3.5 to reformulate the gathered information for each financial term into a concise, paragraph-long textual definition. 
Since some financial terms come with mathematical definitions, we address the issue of varied formula formats in the original sources (e.g., LaTeX and HTML). We instruct GPT-4 to transform these formulas into a unified python program format. \autoref{fig:knowledge_example} illustrates an example of knowledge terms collected in the knowledge bank. 

\paragraph{Knowledge Bank Update and Maintenance}
After formulating knowledge using LLMs, during the dataset annotation stage (Section~\ref{sec:data_annotation}), we dynamically update and maintain the constructed knowledge bank, incorporating new knowledge that, although not initially covered, is essential for answering the annotated questions. Additionally, we remove any duplicate entries identified by the annotators. We eventually collect \nknowledge pieces of financial knowledge in the knowledge bank, with 57.4\% of the terms including Python-formatted mathematical definitions. 

\subsection{\ours Question Annotation}\label{sec:data_annotation}
For each financial term in the knowledge bank, we instruct annotators to create a corresponding math reasoning question, if applicable. The answer to the composed question should be a numeric value. The annotators are required to adhere to the following guidelines for a successful question annotation:

\paragraph{Question Annotation}\label{sec:question-annotation}
If the annotators choose to adapt questions from textbooks or the Internet instead of creating their own from scratch, they are asked to adhere to copyright and license regulations, avoiding data from sites prohibiting copy and redistribution. Furthermore, they are required not only to modify the surface-level description of the question but also to change the associated numeric values. 
In light of the emerging concerns about \emph{data contamination} in LLMs~\cite{shi2023detecting, deng-etal-2024-investigating}, we instruct annotators to conduct a Google search for each annotated question, ensuring that no similar question appears on the first page of the search results. 
Additionally, we recognize that many financial problems involve tables, as shown in \autoref{fig:example}. Such tabular data plays a crucial role in thoroughly understanding financial problems, and it presents unique challenges for LLMs in terms of comprehension and interpretation. Therefore, we encourage and reward annotators to include tables that are relevant and accurately represent the data pertinent to the questions.
Finally, out of \nexample questions, 674 are marked as having been adapted from existing resources, and 482 are accompanied with tabular data.

\paragraph{Identifying Question-relevant Knowledge}~\label{sec:label_knowledge}
After a question is annotated, annotators must identify 1-3 key financial concepts for answering this question. They then search for each term in our constructed knowledge bank. If the term is included, they verify its context and details for relevance. If a term is absent or with low-quality definition, annotators receive a bonus for documenting the term, providing a brief explanation or definition and outlining its relevance to the problem. These identified terms are subsequently added or updated in the knowledge bank, resulting in a total of 123 new inclusions and 47 revisions.

\subsection{\ours Solution Annotation}
As illustrated in \autoref{tab:dataset_comparison}, existing math reasoning benchmarks typically represent solutions using text or mathematical equations. However, solutions in text format often lack the precision and unambiguous nature required for computational problem-solving. Solutions in mathematical equations are explicit, but less descriptive, as the semantic meaning associated with each numeric value in the equations can be ambiguous.
Moreover, these two formats are less adaptable for use in automated systems due to variations in language and difficulties in semantic parsing and execution.

To overcome these limitations, we use Python programs,
starting with ``\texttt{def solution():}'', to represent solutions. Such Python program combines the explicitness of code execution with the descriptive power of annotated comments, offering a more effective and adaptable solution representation for complex math reasoning problems.
Specifically, annotators are required to first define variables with meaningful names at the beginning of the Python function. These variables correspond to the key elements or quantities mentioned in the textual or tabular content of questions. The annotators then proceed to write a sequence of Python statements that logically solve the problem, step by step. 
To ensure the accuracy and functionality of the Python-format solutions, our annotation interface automatically executes the Python function. This execution checks that the return type of the answer is either a float or an int and verifies that there are no execution errors. 

\subsection{Data Quality Validation}
We conduct a comprehensive validation protocol to ensure the high quality of \ours. For each example, we first assign another annotator to validate whether: 1) the question is meaningful and grammatically correct, 2) the associated knowledge terms are accurately annotated and complete, 3) the Python-format solution is logically correct and easy to understand.
Validators are asked to revise examples that do not meet these standards.

We also report the human evaluation scores over 200 randomlysampled examples. As illustrated in \autoref{tab:annotation_aggrement} in the Appendix, \ours has a high annotation quality.

\begin{table}[t!]
\centering 
\resizebox{\columnwidth}{!}{
\addtolength{\tabcolsep}{-0.5em}
\begin{tabular}{lr}
\toprule
\textbf{Property}                            & \textbf{Value}  \\
\midrule
\multicolumn{2}{c}{\textbf{Knowledge Bank}}\\
\noalign{\vskip 1ex}
\# Knowledge Terms & \nknowledge \\
Textual Definition Length \texttt{(Median/Avg)} & 47.1 / 49.7\\
\% \emph{w.} Mathematical Definition & 57.4\%\\
\midrule

\multicolumn{2}{c}{\textbf{\ours Dataset}}\\
\noalign{\vskip 1ex}
Question Length \texttt{(Median/Avg)}   & 54.0 / 61.8            \\
\noalign{\vskip 0.5ex}\hdashline\noalign{\vskip 0.5ex}
\% Questions with Table    & 482 (40.2\%)   \\
\# Rows per Table \texttt{(Median/Avg)} & 3.0 / 3.0 \\
\# Columns per Table \texttt{(Median/Avg)} & 4.0 / 5.0\\
\noalign{\vskip 0.5ex}\hdashline\noalign{\vskip 0.5ex}
\# Knowledge Terms per Example \texttt{(Median/Avg)} & 2.5 / 2.4 \\
\# Math Operations in Python Solution \texttt{(Median/Avg)} & 5.0 / 5.6\\
\# Lines in Python Solution \texttt{(Median/Avg)} & 6.0 / 6.6 \\
\noalign{\vskip 0.5ex}\hdashline\noalign{\vskip 0.5ex}
Development Set Size &  200\\
Test Set Size & 1,000 \\
\bottomrule
\end{tabular}
}
\caption{Basic statistics of the constructed knowledge bank and \ours dataset.}
\label{tab:basic-stats}
\end{table}
\begin{figure}[!t]
    \centering
    \includegraphics[width = 1\linewidth]{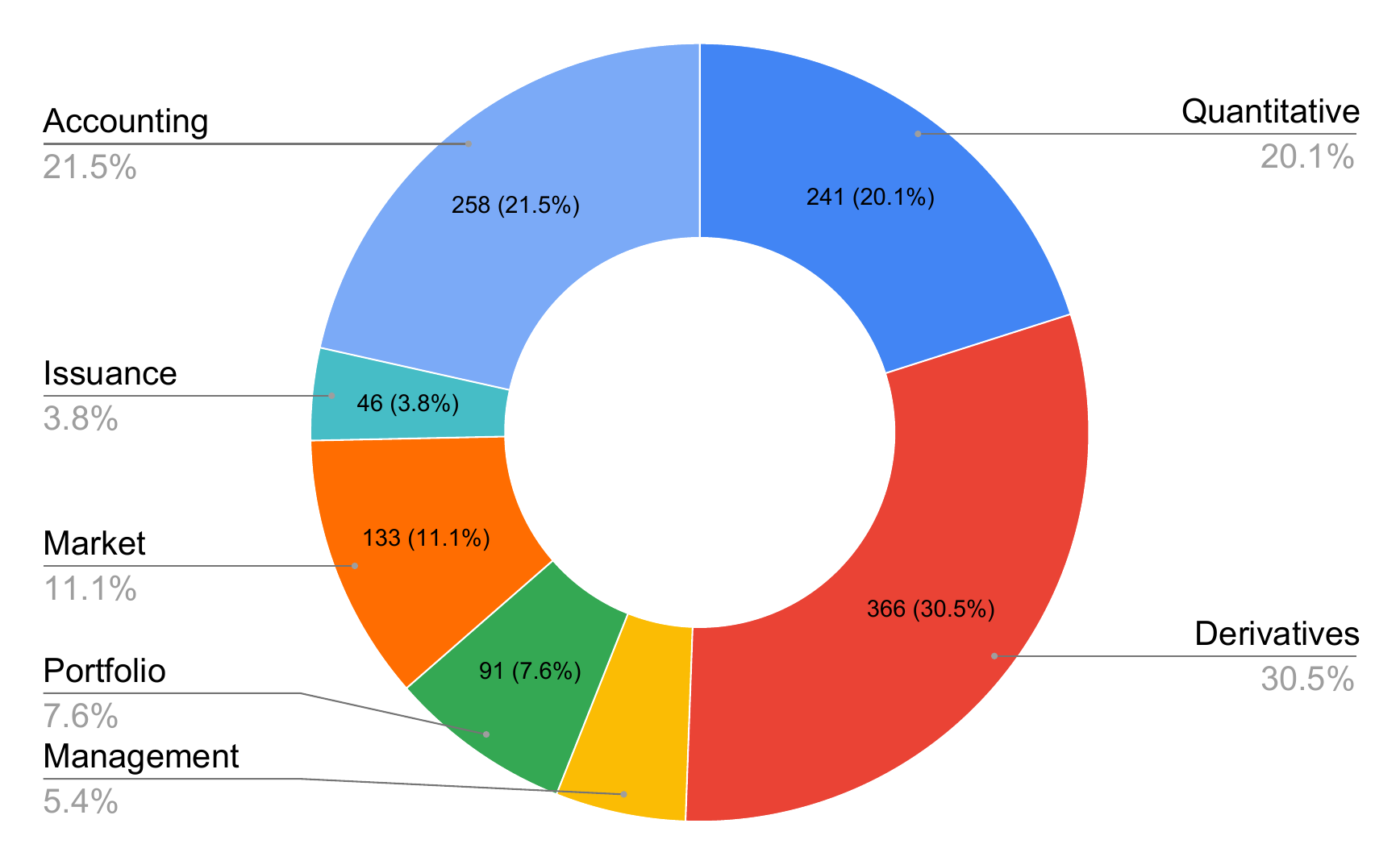}
    \caption{Topic distribution of \ours.}
    \label{fig:topic}
\end{figure}
\subsection{Data Statistics and Dataset Release}\label{sec:data-release}
\autoref{tab:basic-stats} describes the basic statistics of \ours, with topic-type distribution shown in \autoref{fig:topic}.
We randomly divide the dataset into two subsets: \dev and \emph{test}. The \dev set contains 200 examples and is intended for model development validation. The \emph{test} set comprises the remaining 1,000 examples and is designed for standard evaluation. To prevent data contamination~\cite{shi2023detecting, sainz-etal-2023-nlp, deng2024investigating}, the answer for the \emph{test} set will not be publicly released. Instead, we develop and maintain an online evaluation platform, allowing researchers to evaluate models and participate in a leaderboard.
Following recent LLM reasoning benchmarks~\cite{chen-etal-2023-theoremqa, yue2023mmmu, lu2023mathvista}, the main evaluation of \ours is conducted under a \emph{zero-shot} setting on the \emph{test} set to assess LLMs' capabilities to generate accurate answers without fine-tuning or few-shot demonstrations on our benchmark. 

\subsection{Human-level Performance Evaluation}\label{sec:human-performance}
To provide a rough but informative estimate of human-level performance by non-experts and experts on \ours, we randomly sampled 50 examples from the \emph{validation} set. We enroll two experts, both with the CFA license, and two non-experts to individually solve these questions.

We first evaluate their performance in a \emph{closed-book} setting, where the evaluators do not have access to the internet or textbooks and are required to finish the 50 questions within three hours. The non-expert evaluators achieve accuracy of 54\% and 62\% (average 58\%), and the expert evaluators achieve accuracy of 76\% and 70\% (average 73\%).

We then transition to an \emph{open-book} setting, where the evaluators are asked to use the internet and textbooks to correct their initial errors. This setting is designed to assess how external knowledge resources could enhance human problem-solving abilities and accuracy. The non-expert evaluators improved their accuracy to 86\% and 82\% (average 84\%). Similarly, the expert evaluators improved the accuracy to 94\% and 90\% (average 92\%). 

\section{Evaluated Systems}
This section discusses the investigated LLMs and prompting methods in our work.
\subsection{Large Language Models}
We evaluate following LLMs on \ours: 
\begin{itemize} [leftmargin=*]
\itemsep0em 
\item \textbf{General}: 
GPT-3.5\&4~\cite{openai2022chatgpt, OpenAI2023GPT4TR, openai2024gpt4o},
Gemini-1.5~\cite{geminiteam2024gemini},
Claude-3\&3.5~\cite{claude3},
Llama-2\&3\&3.1~\cite{Touvron2023Llama2O, dubey2024llama3}, 
Mistral~\cite{jiang2023mistral}, 
Phi-3~\cite{abdin2024phi3}, 
Gemma-1\&2~\cite{gemmateam2024gemma1},
WizardLM-2~\cite{xu2023wizardlm}, 
Yi-1.5~\cite{yi}, 
Qwen-2~\cite{qwen}, 
Command R+~\cite{commandr},
Aya~\cite{aya},
and GLM-4~\cite{glm2024chatglm}.

\item \textbf{Math-specific}:
WizardMath~\cite{luo2023wizardmath}, 
DeepSeek-Math~\cite{shao2024deepseekmath},
Mathtral~\cite{mathstralmodelcard},
and InternLM-Math~\cite{ying2024internlmmath}.

\item \textbf{Code-based}: 
DeepSeek-Coder-V1~\cite{guo2024deepseekcoder},
WizardCoder~\cite{luo2023wizardcoder}, 
Codestral~\cite{codestral},
DeepSeek-Coder-V2 (also MoE architecture, ~\citet{deepseekai2024deepseekv2}),
and StarCoder2~\cite{lozhkov2024starcoder2}.

\item \textbf{Mixture of Experts (MoE)}: 
Mixtral~\cite{mistral2023moe},
WizardLM-2 (MoE, ~\citet{xu2023wizardlm}), 
DeepSeek-V2~\cite{deepseekai2024deepseekv2},
and DBRX~\cite{dbrx}.

\end{itemize}

We select the most recent checkpoint available as of August 1, 2024. 
The details of each evaluated model, including the exact model version we used, are presented in \autoref{tab:model-detail} in Appendix. The experiments for open-sourced LLMs were conducted using \texttt{vLLM} framework~\cite{kwon2023efficient}. 
For all the experiments, we set temperature as 1.0, Top P as 1.0, and maximum output length as 512.

\begin{figure}[!t]
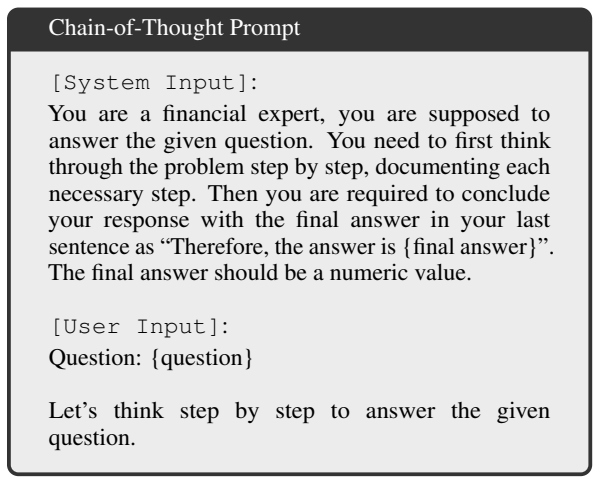

\begin{tcolorbox}[colback=black!7.5!white, colframe=black!80!white, title=Chain-of-Thought Prompt, fontupper=\footnotesize, fonttitle=\footnotesize]
\texttt{[System Input]}: \vspace{2pt}\\
You are a financial expert, you are supposed to answer the given question. You need to first think through the problem step by step, documenting each necessary step. Then you are required to conclude your response with the final answer in your last sentence as ``Therefore, the answer is \{final answer\}''. The final answer should be a numeric value. \\
\newline
\texttt{[User Input]}: \vspace{2pt}\\
Question: \{question\}\\
\newline
Let's think step by step to answer the given question.
\end{tcolorbox}

\caption{Example of \emph{zero}-shot CoT prompt used.}
\label{fig:cot}
\end{figure}
\subsection{Prompting Methods}
Following \citet{chen-etal-2023-theoremqa} and \citet{lu2023mathvista}, we evaluate two established prompting methods, with examples of prompt illustrated in \autoref{fig:cot} and \autoref{fig:pot} in the Appendix, respectively.

\paragraph{Chain-of-Thought} The CoT method~\cite{wei2022chain, kojima2022large} instructs the LLMs to articulate a step-by-step reasoning process. This leads to a detailed explanation that culminates in the final answer. 

\paragraph{Program-of-Thought} Different from CoT, the PoT method~\cite{chen2023program} disentangles computation from the reasoning process by prompting the LLMs to generate a structured program to represent the reasoning process. The final answer is then derived by executing the generated program with an external calculator.
\begin{table*}[!t]
\centering
\resizebox{0.94\textwidth}{!}{%
\renewcommand{\arraystretch}{1.1}
\addtolength{\tabcolsep}{-0.1em}
\begin{tabular}{lrl:rrrrrrrrrrrrrrrrrrrrr:rr}
\toprule
\multirow{2}{*}{\textbf{Model}} & \multirow{2}{*}{\textbf{Size}}  & \multirow{2}{*}{\textbf{Notes}} & \multicolumn{2}{c}{\textbf{Quantitative}} && \multicolumn{2}{c}{\textbf{Derivatives}} && \multicolumn{2}{c}{\textbf{Accounting}} && \multicolumn{2}{c}{\textbf{Management}} && \multicolumn{2}{c}{\textbf{Portfolio}} && \multicolumn{2}{c}{\textbf{Economics}} && \multicolumn{2}{c}{\textbf{Corporate}} && \multicolumn{2}{c}{\textbf{Avg.}} \\
\cmidrule(lr){4-5} \cmidrule(lr){7-8} \cmidrule(lr){10-11} \cmidrule(lr){13-14} \cmidrule(lr){16-17} \cmidrule(lr){19-20} \cmidrule(lr){22-23} \cmidrule(lr){25-26}
& & & PoT & CoT && PoT & CoT && PoT & CoT && PoT & CoT &&  PoT & CoT  &&PoT & CoT &&PoT & CoT && PoT & \textbf{CoT}\\
\midrule
Close-book \\
\quad Expert &  & &  & & & &   &  &  &  &  &  &  &  &  &  &  &  &  &  &  &  &  & \multicolumn{2}{c}{73.0} \\
\quad Non-Expert &  & & & &  &  &  &  &  &  &  &  &  &  &  &  &  &  &  &  &  &  &  & \multicolumn{2}{c}{58.0} \\
\noalign{\vskip 0.5ex}\hdashline\noalign{\vskip 0.5ex}
Open-book \\
\quad Expert &  & &  & & & &   &  &  &  &  &  &  &  &  &  &  &  &  &  &  &  &  & \multicolumn{2}{c}{92.0} \\
\quad Non-Expert &  & & & & &   &  &  &  &  &  &  &  &  &  &  &  &  &  &  &  &  &  & \multicolumn{2}{c}{84.0} \\

\midrule
\multicolumn{26}{c}{\emph{\textbf{Proprietary LLMs}}} \\\noalign{\vskip 0.5ex}

GPT-4o &  &  &\cellcolor{blue!35}75.0 & 45.8 &  & \cellcolor{blue!35}58.8 & \cellcolor{blue!35}55.4 &  & \cellcolor{blue!5}60.3 & \cellcolor{blue!20}69.4 &  & \cellcolor{blue!35}82.9 & \cellcolor{blue!35}66.3 &  & \cellcolor{blue!35}77.8 & \cellcolor{blue!5}69.4 &  & \cellcolor{blue!20}62.5 & \cellcolor{blue!35}58.9 &  & \cellcolor{blue!35}67.0 & \cellcolor{blue!20}56.9 &  & \cellcolor{blue!35}\underline{67.0} & \cellcolor{blue!35}60.9 \\
Claude-3.5-Sonnet &  &  &\cellcolor{blue!20}73.6 & \cellcolor{blue!20}55.6 &  & \cellcolor{blue!20}54.1 & \cellcolor{blue!20}51.8 &  & \cellcolor{blue!35}66.7 & \cellcolor{blue!35}69.9 &  & \cellcolor{blue!5}75.6 & \cellcolor{blue!5}63.9 &  & \cellcolor{blue!5}72.2 & \cellcolor{blue!20}72.2 &  & \cellcolor{blue!35}64.3 & \cellcolor{blue!20}58.9 &  & \cellcolor{blue!20}62.4 & \cellcolor{blue!35}60.6 &  & \cellcolor{blue!20}\underline{64.8} & \cellcolor{blue!20}60.6 \\
Claude-3-Opus &  &  &66.7 & \cellcolor{blue!35}56.9 &  & \cellcolor{blue!5}53.5 & \cellcolor{blue!5}45.2 &  & \cellcolor{blue!20}62.1 & 59.8 &  & \cellcolor{blue!20}79.5 & \cellcolor{blue!20}64.9 &  & \cellcolor{blue!20}72.2 & \cellcolor{blue!35}83.3 &  & 51.8 & 46.4 &  & \cellcolor{blue!5}59.6 & 45.0 &  & \cellcolor{blue!5}\underline{62.9} & \cellcolor{blue!5}54.7 \\
GPT-4-Turbo &  &  &59.7 & 38.9 &  & 49.8 & 42.2 &  & 50.7 & \cellcolor{blue!5}64.8 &  & 72.2 & 56.6 &  & 61.1 & 50.0 &  & 57.1 & 44.6 &  & 50.5 & \cellcolor{blue!5}47.7 &  & \underline{56.2} & 50.9 \\
Gemini-1.5-Pro &  &  &68.1 & \cellcolor{blue!5}50.0 &  & 53.1 & 30.7 &  & 56.6 & 55.2 &  & 69.8 & 57.6 &  & 58.3 & 63.9 &  & 51.8 & \cellcolor{blue!5}55.4 &  & 50.5 & 44.0 &  & \underline{58.2} & 47.0 \\
GPT-4o-Mini &  &  &65.3 & 36.1 &  & 46.9 & 29.7 &  & 48.4 & 47.5 &  & 69.3 & 46.8 &  & 50.0 & 38.9 &  & \cellcolor{blue!5}57.1 & 41.1 &  & 51.4 & 45.9 &  & \underline{54.3} & 40.3 \\
Gemini-1.5-Flash &  &  &\cellcolor{blue!5}69.4 & 33.3 &  & 43.6 & 28.7 &  & 52.0 & 48.9 &  & 67.8 & 49.8 &  & 58.3 & 61.1 &  & 50.0 & 37.5 &  & 47.7 & 34.9 &  & \underline{53.6} & 40.1 \\
Claude-3-Sonnet &  &  &59.7 & 37.5 &  & 37.0 & 28.4 &  & 48.0 & 43.4 &  & 66.8 & 48.8 &  & 47.2 & 55.6 &  & 48.2 & 33.9 &  & 48.6 & 35.8 &  & \underline{49.4} & 38.6 \\
Claude-3-Haiku &  &  &34.7 & 31.9 &  & 19.8 & 26.4 &  & 33.8 & 43.4 &  & 44.9 & 41.5 &  & 41.7 & 44.4 &  & 25.0 & 33.9 &  & 36.7 & 30.3 &  & 32.0 & 35.1 \\
GPT-3.5-Turbo &  &  &47.2 & 25.0 &  & 24.4 & 16.5 &  & 29.2 & 29.2 &  & 51.2 & 33.2 &  & 27.8 & 22.2 &  & 37.5 & 21.4 &  & 32.1 & 23.8 &  & \underline{34.3} & 24.6 \\

\midrule
\multicolumn{26}{c}{\emph{\textbf{Open-source LLMs}}} \\\noalign{\vskip 0.5ex}

DeepSeek-V2 & 236B & MoE &\cellcolor{blue!20}72.2 & \cellcolor{blue!5}43.1 &  & \cellcolor{blue!20}49.8 & \cellcolor{blue!35}46.5 &  & \cellcolor{blue!5}56.6 & \cellcolor{blue!5}53.9 &  & \cellcolor{blue!35}77.6 & \cellcolor{blue!35}68.8 &  & 61.1 & \cellcolor{blue!20}63.9 &  & \cellcolor{blue!35}57.1 & \cellcolor{blue!5}44.6 &  & \cellcolor{blue!35}59.6 & \cellcolor{blue!35}56.9 &  & \cellcolor{blue!35}\underline{60.5} & \cellcolor{blue!35}54.1 \\
DeepSeek-Coder-V2 & 236B & Code &\cellcolor{blue!5}65.3 & \cellcolor{blue!20}44.4 &  & \cellcolor{blue!35}50.2 & \cellcolor{blue!20}43.6 &  & \cellcolor{blue!35}58.9 & \cellcolor{blue!35}56.6 &  & \cellcolor{blue!20}77.1 & \cellcolor{blue!20}67.3 &  & 52.8 & \cellcolor{blue!35}72.2 &  & 51.8 & \cellcolor{blue!35}50.0 &  & \cellcolor{blue!20}57.8 & \cellcolor{blue!20}53.2 &  & \cellcolor{blue!5}\underline{59.7} & \cellcolor{blue!20}53.8 \\
Llama-3.1 & 405B &  &\cellcolor{blue!35}76.4 & \cellcolor{blue!35}50.0 &  & \cellcolor{blue!5}48.8 & \cellcolor{blue!5}34.6 &  & \cellcolor{blue!20}58.0 & \cellcolor{blue!20}56.6 &  & \cellcolor{blue!5}76.6 & \cellcolor{blue!5}52.7 &  & \cellcolor{blue!35}69.4 & 55.6 &  & \cellcolor{blue!20}57.1 & 41.1 &  & \cellcolor{blue!5}54.1 & \cellcolor{blue!5}47.7 &  & \cellcolor{blue!20}\underline{60.3} & \cellcolor{blue!5}46.8 \\
Llama-3.1 & 70B &  &62.5 & 38.9 &  & 39.3 & 29.7 &  & 47.5 & 53.0 &  & 65.8 & 50.7 &  & \cellcolor{blue!5}61.1 & \cellcolor{blue!5}58.3 &  & 44.6 & \cellcolor{blue!20}44.6 &  & 44.0 & 36.7 &  & \underline{49.8} & 42.4 \\
Mistral-Large & 123B &  &59.7 & 36.1 &  & 44.9 & 29.4 &  & 49.8 & 48.0 &  & 75.6 & 47.3 &  & \cellcolor{blue!20}63.9 & 36.1 &  & 50.0 & 33.9 &  & 52.3 & 44.0 &  & \underline{55.1} & 39.7 \\
Qwen2 & 72B &  &41.7 & 30.6 &  & 30.7 & 23.4 &  & 38.8 & 48.0 &  & 52.7 & 42.0 &  & 44.4 & 50.0 &  & 39.3 & 33.9 &  & 38.5 & 36.7 &  & \underline{39.6} & 36.1 \\
Llama-3 & 70B &  &56.9 & 36.1 &  & 39.9 & 23.4 &  & 46.6 & 43.4 &  & 65.8 & 42.9 &  & 58.3 & 47.2 &  & \cellcolor{blue!5}51.8 & 39.3 &  & 44.0 & 34.9 &  & \underline{49.7} & 35.7 \\
Phi-3-Medium & 14B &  &40.3 & 31.9 &  & 31.4 & 23.1 &  & 36.1 & 47.0 &  & 54.2 & 42.4 &  & 41.7 & 50.0 &  & 39.3 & 32.1 &  & 35.8 & 28.4 &  & \underline{39.0} & 35.0 \\
Mixtral-8x22B & 141B & MoE &15.3 & 31.9 &  & 4.6 & 23.4 &  & 9.6 & 35.2 &  & 22.9 & 38.0 &  & 5.6 & 36.1 &  & 16.1 & 33.9 &  & 3.7 & 30.3 &  & 10.8 & 31.4 \\
DeepSeek-Coder-V2-Lite & 16B & Code &38.9 & 36.1 &  & 19.1 & 21.1 &  & 23.3 & 31.5 &  & 49.3 & 42.4 &  & 33.3 & 30.6 &  & 26.8 & 26.8 &  & 26.6 & 26.6 &  & 29.4 & 30.1 \\
Gemma-2 & 9B &  &30.6 & 34.7 &  & 19.1 & 19.8 &  & 31.5 & 34.7 &  & 43.4 & 37.6 &  & 25.0 & 33.3 &  & 33.9 & 25.0 &  & 27.5 & 26.6 &  & \underline{29.6} & 29.3 \\
Yi-1.5 & 9B &  &23.6 & 29.2 &  & 14.8 & 14.8 &  & 18.3 & 35.2 &  & 20.0 & 40.0 &  & 16.7 & 27.8 &  & 10.7 & 25.0 &  & 18.4 & 30.3 &  & 17.5 & 28.2 \\
Yi-1.5 & 34B &  &19.4 & 25.0 &  & 16.2 & 16.8 &  & 18.3 & 35.2 &  & 23.9 & 36.6 &  & 19.4 & 19.4 &  & 23.2 & 28.6 &  & 18.4 & 28.4 &  & 19.2 & 27.5 \\
WizardLM-2 & 141B & MoE &30.6 & 23.6 &  & 18.5 & 17.2 &  & 26.0 & 33.3 &  & 40.0 & 32.2 &  & 30.6 & 30.6 &  & 25.0 & 33.9 &  & 25.7 & 29.4 &  & 27.0 & 27.0 \\
Phi-3-Mini & 3B &  &27.8 & 22.2 &  & 12.9 & 13.5 &  & 28.3 & 31.0 &  & 38.0 & 35.1 &  & 27.8 & 19.4 &  & 28.6 & 32.1 &  & 16.5 & 20.2 &  & 24.3 & 24.4 \\
Mistral-Nemo & 12B &  &31.9 & 25.0 &  & 13.5 & 13.2 &  & 12.8 & 25.1 &  & 32.7 & 31.2 &  & 16.7 & 27.8 &  & 14.3 & 23.2 &  & 19.3 & 24.8 &  & 19.4 & 22.7 \\
DBRX & 132B & MoE &12.5 & 20.8 &  & 11.2 & 14.5 &  & 13.2 & 27.4 &  & 26.3 & 27.3 &  & 11.1 & 22.2 &  & 16.1 & 28.6 &  & 17.4 & 21.1 &  & 15.8 & 22.2 \\
DeepSeek-Math & 7B & Math &0.0 & 22.2 &  & 0.3 & 12.2 &  & 0.9 & 19.6 &  & 2.4 & 30.7 &  & 0.0 & 30.6 &  & 1.8 & 26.8 &  & 0.9 & 22.9 &  & 1.0 & 21.0 \\
Qwen2 & 7B &  &8.3 & 18.1 &  & 5.3 & 12.9 &  & 4.6 & 24.7 &  & 19.5 & 29.3 &  & 5.6 & 16.7 &  & 14.3 & 23.2 &  & 6.4 & 18.4 &  & 8.9 & 20.5 \\
GLM-4 & 9B &  &27.8 & 20.8 &  & 13.2 & 11.6 &  & 25.1 & 25.6 &  & 36.1 & 23.4 &  & 22.2 & 25.0 &  & 19.6 & 28.6 &  & 21.1 & 19.3 &  & \underline{23.1} & 20.0 \\
C4AI Command R+ & 104B &  &2.8 & 22.2 &  & 1.3 & 10.2 &  & 1.4 & 23.3 &  & 5.4 & 22.9 &  & 2.8 & 25.0 &  & 0.0 & 14.3 &  & 1.8 & 22.9 &  & 2.3 & 18.7 \\
Mixtral-8x7B-v0.1 & 46B & MoE &0.0 & 22.2 &  & 0.0 & 11.9 &  & 0.0 & 16.4 &  & 1.5 & 24.9 &  & 0.0 & 13.9 &  & 0.0 & 17.9 &  & 0.0 & 21.1 &  & 0.3 & 17.7 \\
Llama-3.1 & 8B &  &22.2 & 18.1 &  & 12.9 & 10.2 &  & 14.6 & 14.2 &  & 35.1 & 28.3 &  & 19.4 & 16.7 &  & 21.4 & 19.6 &  & 16.5 & 22.0 &  & \underline{19.6} & 17.4 \\
Mathstral & 7B & Math &18.1 & 18.1 &  & 7.9 & 9.6 &  & 11.0 & 18.3 &  & 26.8 & 24.9 &  & 13.9 & 8.3 &  & 19.6 & 19.6 &  & 14.7 & 16.5 &  & 14.8 & 16.5 \\
Codestral & 22B & Code &41.7 & 13.9 &  & 16.8 & 9.9 &  & 23.7 & 16.0 &  & 54.2 & 25.8 &  & 19.4 & 11.1 &  & 42.9 & 16.1 &  & 26.6 & 19.3 &  & \underline{30.4} & 16.2 \\
Llama-3 & 8B &  &13.9 & 11.1 &  & 12.2 & 7.6 &  & 17.8 & 18.3 &  & 26.3 & 20.0 &  & 16.7 & 16.7 &  & 17.9 & 14.3 &  & 19.3 & 15.6 &  & \underline{17.7} & 14.3 \\
WizardLM-2 & 7B &  &23.6 & 8.3 &  & 6.6 & 6.9 &  & 11.4 & 18.7 &  & 22.0 & 19.0 &  & 11.1 & 8.3 &  & 16.1 & 8.9 &  & 12.8 & 14.7 &  & \underline{13.4} & 13.1 \\
WizardMath & 7B & Math &5.6 & 11.1 &  & 5.0 & 6.3 &  & 10.5 & 16.9 &  & 18.5 & 12.7 &  & 5.6 & 11.1 &  & 7.1 & 21.4 &  & 9.2 & 15.6 &  & 9.6 & 12.3 \\
DeepSeek-V2-Lite & 16B & MoE &5.6 & 13.9 &  & 1.0 & 6.3 &  & 2.7 & 14.2 &  & 7.3 & 16.1 &  & 2.8 & 11.1 &  & 3.6 & 8.9 &  & 3.7 & 15.6 &  & 3.5 & 11.9 \\
Mistral-v0.3 & 7B &  &1.4 & 13.9 &  & 1.3 & 4.6 &  & 1.4 & 15.1 &  & 6.8 & 15.6 &  & 2.8 & 8.3 &  & 0.0 & 10.7 &  & 2.8 & 11.9 &  & 2.6 & 11.1 \\
Aya-23 & 35B &  &0.0 & 8.3 &  & 0.3 & 7.9 &  & 0.0 & 13.7 &  & 1.0 & 12.7 &  & 0.0 & 11.1 &  & 0.0 & 10.7 &  & 0.0 & 10.1 &  & 0.3 & 10.7 \\
InternLM2-Math-Plus & 7B & Math &8.3 & 16.7 &  & 3.0 & 4.6 &  & 5.9 & 9.1 &  & 14.2 & 19.5 &  & 0.0 & 8.3 &  & 12.5 & 12.5 &  & 10.1 & 8.3 &  & 7.5 & 10.5 \\
Llama-2 & 70B &  &13.9 & 8.3 &  & 3.6 & 6.6 &  & 14.2 & 14.2 &  & 12.2 & 12.2 &  & 5.6 & 13.9 &  & 8.9 & 5.4 &  & 10.1 & 11.9 &  & 9.5 & 10.3 \\
InternLM2 & 7B &  &5.6 & 6.9 &  & 4.6 & 4.0 &  & 6.4 & 11.9 &  & 16.1 & 15.6 &  & 5.6 & 2.8 &  & 12.5 & 7.1 &  & 5.5 & 10.1 &  & 8.0 & 9.1 \\
StarCoder2 & 15B & Code &29.2 & 2.8 &  & 12.5 & 4.3 &  & 11.9 & 9.6 &  & 35.6 & 15.6 &  & 11.1 & 2.8 &  & 16.1 & 12.5 &  & 20.2 & 8.3 &  & \underline{19.3} & 8.5 \\
Gemma-1 & 7B &  &2.8 & 5.6 &  & 1.0 & 3.6 &  & 1.8 & 10.5 &  & 2.9 & 7.8 &  & 0.0 & 5.6 &  & 1.8 & 7.1 &  & 4.6 & 11.0 &  & 2.1 & 7.2 \\
WizardCoder & 33B & Code &19.4 & 4.2 &  & 5.0 & 2.6 &  & 6.8 & 5.5 &  & 37.1 & 10.7 &  & 8.3 & 5.6 &  & 21.4 & 3.6 &  & 11.9 & 9.2 &  & \underline{14.8} & 5.9 \\
DeepSeek-Coder-V1 & 33B & Code &12.5 & 4.2 &  & 2.0 & 3.0 &  & 6.4 & 5.0 &  & 15.6 & 8.3 &  & 5.6 & 5.6 &  & 10.7 & 7.1 &  & 8.3 & 5.5 &  & \underline{7.8} & 5.2 \\
Llama-2 & 7B &  &4.2 & 0.0 &  & 1.0 & 1.6 &  & 2.3 & 5.5 &  & 2.9 & 5.4 &  & 5.6 & 11.1 &  & 0.0 & 3.6 &  & 2.8 & 9.2 &  & 2.2 & 4.4 \\
Aya-23 & 8B &  &1.4 & 1.4 &  & 0.3 & 2.3 &  & 0.0 & 4.6 &  & 0.0 & 5.8 &  & 0.0 & 2.8 &  & 0.0 & 0.0 &  & 0.0 & 8.3 &  & 0.2 & 4.0 \\
Gemma-1 & 2B &  &0.0 & 0.0 &  & 1.3 & 2.0 &  & 3.2 & 6.4 &  & 5.8 & 5.4 &  & 8.3 & 0.0 &  & 0.0 & 0.0 &  & 1.8 & 5.5 &  & 2.8 & 3.7 \\

\bottomrule
\end{tabular}
}
\caption{Results of Chain-of-Thought and Program-of-Thought prompting on the \emph{test} set of \ours.
% We select the most recent version as of July 5, 2024, for each model.
We use average Accuracy using CoT prompting as the ranking indicator of model performance. \underline{Numbers} underscored indicate that models with PoT prompting achieves better results than with CoT prompting. }
% Models marked with $^*$ are found challenging to instruct for generating \emph{Python-format solutions} using PoT prompting; therefore, we only report their performance with CoT prompting approach.}
\label{tab:results}
\end{table*}
\section{Experiments}
\subsection{Experiment Setup}
\paragraph{Final Answer Extraction}
For LLM with CoT prompting, we adopt the answer extraction pipeline from \citet{chen-etal-2023-theoremqa} to identify the final answer from the model's output. For LLM with PoT prompting, we first extract the generated python solution from the model's output. If this python solution is executable, we execute it to obtain the final answer. Once we obtain the final answer from model's output, we compare it with the ground-truth answer for accuracy measurement.

\paragraph{Tabular Data Serialization}
Following previous work on table-relevant tasks~\cite{chen-2023-large, zhao-etal-2023-investigating}, we use Markdown format to present tabular data in math reasoning problems. In our preliminary study, we discovered that GPT-* and Llama-3 models can effectively understand such table representations.

\subsection{Main Results}
\autoref{tab:results} and \autoref{tab:results-dev} in Appendix illustrate the performance of the evaluated LLMs using CoT and PoT prompting methods on the \ours test and development sets, respectively. 

The experimental results demonstrate that \ours poses significant challenges to current LLMs.
Even the best-performing LLM, GPT-4o, performs much worse than human experts. Specifically, the accuracy of GPT-4o using the CoT prompting method stands at 60.9\%, falling short of the 92\% accuracy achieved by expert evaluators in the open-book setting. This gap highlights the critical need for further advancements in LLMs, especially in complex problem solving within specialized domains that are knowledge-intensive.

Open-source LLMs still significantly lag behind the most advanced versions of the three major families of proprietary LLMs. 
However, the two DeepSeek-V2 models are an exception. They achieve performance levels close to those of the best-performing proprietary models. 
This indicates the potential of open-source LLMs to close the performance gap with proprietary models in the near future, given continued innovation and community collaboration.
Additionally, the proprietary LLMs and code-specific models typically achieve comparable or better performance when using PoT prompting compared to CoT prompting.
For math-specific LLMs, InternLM2-Math-Plus surpasses its backbone in CoT, improving from 9.1\% to 10.5\%. This demonstrates the effectiveness of instruction-tuning in enhancing math reasoning.

\begin{figure}[!t]
    \centering
    \includegraphics[width = \linewidth]{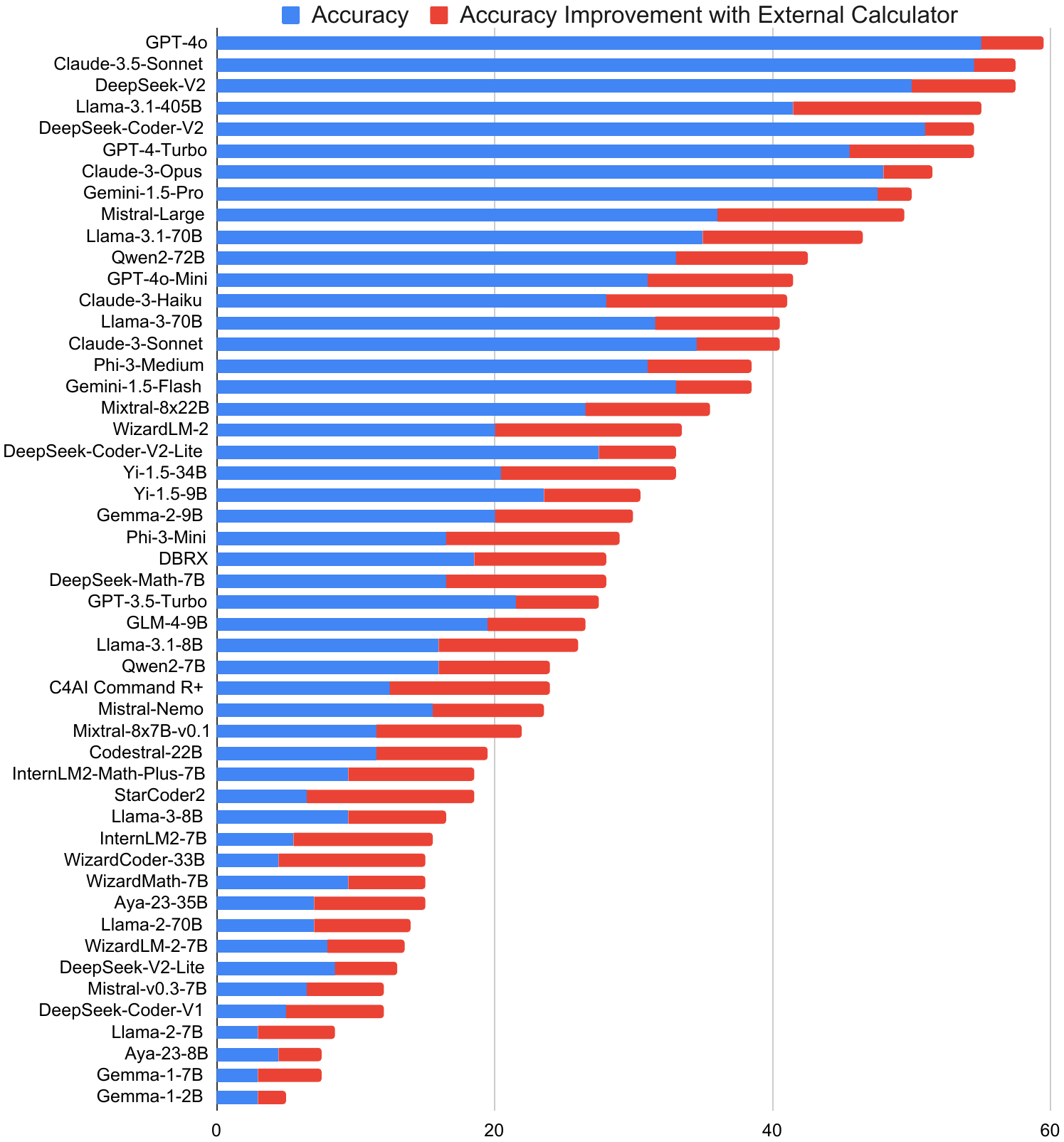}
    \caption{Calibrated results of Chain-of-Thought prompting on the \textbf{\dev} set with an external calculator for math computation. Performing complex math computations correctly is still challenging for LLMs, especially open-source ones.}
    \label{fig:cot_calculator}
\end{figure}
\subsection{Error Analysis}
To gain a deeper insight into the capabilities and limitations of open-source LLMs on our dataset, we conduct a comprehensive error analysis and case studies. 
The error analysis is based on 50 sampled failure cases of Llama-3-70B from the \dev set.
We choose the Llama-3 model as the focus since many open-source models are developed using it as the backbone.
We identify three common mistakes of current LLMs:  \textbf{(1) Misinterpretation of Required Knowledge} (27/50): the model fails to accurately identify and interpret the domain-specific knowledge needed to answer a question correctly, leading to incorrect responses. \autoref{tab:error_analysis} in Appendix illustrates an error example.
\textbf{(2) Incorrect Math Computation} (19/50): the mathematical computation in the intermediate or final step is incorrect, although the reasoning process is correct.
\textbf{(3) Table Misunderstanding} (3/50): The model misinterprets the data within complex-structure tables.

To separate computational abilities from final accuracy, we employed an external calculator~\cite{inaba-etal-2023-multitool} for CoT outputs. Specifically, we used GPT-3.5-Turbo to extract single-line math expressions from the models' textual responses and executed these expressions to obtain the final answers. \autoref{fig:cot_calculator} illustrates the calibrated results of LLM CoT performance with an external calculator. It demonstrates that performing complex math computations correctly is still challenging for LLMs, especially open-source ones.

\subsection{Program-of-Thought Analysis}
To better analyze the PoT prompting methods, we examine the execution rate of each LLM under PoT prompting, measuring how many of the generated Python programs are executable. 
\autoref{fig:execution-rate} in the Appendix illustrates the relationship between execution rate and accuracy across different models. 
It demonstrates that for models unable to consistently generate executable programs (\ie models with an execution rate < 60\%), their degraded performance when applying PoT prompting is attributable to the low execution rate.
For instance, although Mistral-8$\times$22B achieves competitive performance with CoT, it struggles to consistently generate executable Python solutions, leading to lower accuracy with the PoT prompting approach.
Conversely, for LLMs capable of generating executable programs (i.e., models with an execution rate > 80\%), the final answer accuracy is mainly attributed to the reasoning capabilities of the models.

\section{Knowledge Augmentation Analysis}\label{sec:knowledge-analysis}
In this section, we provide a comprehensive analysis to understand the performance of LLMs and the quality of knowledge incorporated into the input context, aiming to provide insights for future work on solving knowledge-intensive tasks.

\subsection{Evaluated Knowledge-Augmented Method}
We develop and evaluate various knowledge-augmented approaches. For each setting, we include the definition of question-relevant knowledge terms within the prompts (\autoref{fig:knowledge_aug_cot} in Appendix).

\begin{itemize} [leftmargin=*]
\itemsep0em 
\item \textbf{Oracle:}
To investigate the headroom in knowledge augmentation, we use an oracle setting, where the \emph{ground-truth} knowledge terms associated with the question (Section~\ref{sec:question-annotation}) are included.

\item \textbf{LLM as Knowledge Base:}
Recent work~\cite{petroni-etal-2019-language, kang2023knowledgeaugmented} demonstrates that LLMs themselves can effectively serve as knowledge bases. 
This approach is particularly valuable in scenarios where an external knowledge base is unavailable.
We prompt LLMs to first identify the financial terms required to answer the question. They then generate definitions of each identified knowledge term using the inherent data memorization capabilities. 

\item \textbf{Knowledge Retrieval:} We use the question as the retrieval query to the constructed knowledge bank. We investigate 1) BM25 as sparse retriever and 2) OpenAI Text Embedding V3 Large
% \footnote{We use the \texttt{text-embedding-3-large} version (\url{https://platform.openai.com/docs/guides/embeddings}).}
as dense retriever to retrieve the top-$n$ question-relevant knowledge terms from knowledge bank.

\item \textbf{LLM-Instructed Knowledge Retrieval:}
While the method of using ``LLM as Knowledge Base'' can effectively identify the knowledge required to answer a question, it is likely to produce knowledge definitions that are not entirely accurate~\cite{chen2023beyond, peng2023check}. To address this unfaithfulness issue, we harness the power of external knowledge retrieval for obtaining more trustworthy knowledge definitions. Specifically, instead of using the original question as the retrieval query, we utilize each knowledge term along with its definition generated from the ``LLM as Knowledge Base''. This approach provides a more informative and semantically similar basis for knowledge retrieval.

\item \textbf{LLM as Retrieval Re-Ranker:}
Recent studies have demonstrated LLMs' competitive capabilities in re-ranking retrieved candidates to output a more precise list~\cite{sun-etal-2023-chatgpt}. Therefore, in this setting, we first use retriever in ``Knowledge Retrieval'' to retrieve top-$3n$ candidates. Subsequently, we prompt LLMs to select top-$n$ most relevant knowledge terms from this candidate set. 

\end{itemize}

\begin{table}[!t]
    \centering
    \resizebox{\linewidth}{!}{
    \begin{tabular}{lll}
        \toprule
        Setting & Llama-3-70B & Gemini-1.5-Pro\\
        \midrule 
        \emph{wo.} knowledge augmentation & 31.5 & 47.5\\
        \noalign{\vskip 0.5ex}\hdashline\noalign{\vskip 0.5ex}
        
        LLM as Knowledge Base & 29.0 \down{2.5} & 48.5 \up{1.0} \\
        \noalign{\vskip 0.5ex}\hdashline\noalign{\vskip 0.5ex}
        
        BM25 ($n=3$)& \\
        \quad Vanilla Retrieval & 30.0 \down{1.5} & 44.5 \down{3.0}\\
        \quad LLM as Retrieval Re-Ranker & 32.0 \up{0.5}&  49.0 \up{1.5}\\
        \quad LLM-instructed Retrieval & 32.5 \up{1.0} & 48.0 \up{0.5} \\
        
        \noalign{\vskip 0.5ex}\hdashline\noalign{\vskip 0.5ex}
        OpenAI Embedding-3-L ($n=3$)& \\
        \quad Vanilla Retrieval & 32.5 \up{1.0} & 49.0 \up{1.5}\\
        \quad LLM as Retrieval Re-Ranker & 33.5 \up{2.0} & 50.5 \up{3.0}\\
        \quad LLM-instructed Retrieval & 33.5 \up{2.0} & 52.0 \up{4.5}\\
        
        \noalign{\vskip 0.5ex}\hdashline\noalign{\vskip 0.5ex}
        Oracle & 37.5 \up{6.0} & 54.5 \up{7.0}\\
        \bottomrule
    \end{tabular}
    }
    \caption{Results of Chain-of-Thought prompting approach under different knowledge augmentation settings on the \dev set of \ours.}
    \label{tab:retriever_performance}
\end{table}
\subsection{Knowledge Augmentation Results}  
As illustrated in \autoref{tab:retriever_performance}, improving the question-relevance of incorporated knowledge can consistently improve the LLMs' performance. 
Specifically, LLMs equipped with retrieved knowledge from OpenAI Text Embedding consistently outperform those using retrieved knowledge from BM25, due to the more advanced retrieval capabilities of the former.
Among different LLM-aided retrieval strategies, \emph{LLM-Instructed Knowledge Retrieval} achieves the best performance, demonstrating the effectiveness of using \emph{refined} queries for knowledge retrieval.
Nevertheless, it is worth noting that even when incorporated with the ground-truth knowledge (\ie the oracle setting), Gemini-1.5-Pro still performs much worse than human experts in close-book setting (\ie 92.0\%). This highlights the need for future work on developing more advanced domain-specific knowledge integration methods.

\section{Related Work}
The development of general-purpose intelligent systems is significantly dependent on the foundational aspect of mathematical reasoning, a topic that has garnered considerable attention in the academic community. As illustrated in \autoref{tab:dataset_comparison}, researchers have proposed a wide spectrum of math reasoning datasets that cater to a variety of educational levels, ranging from elementary school to college
~\cite{koncel-kedziorski-etal-2016-mawps,wang-etal-2017-deep,amini-etal-2019-mathqa, miao-etal-2020-diverse, patel-etal-2021-nlp,cobbe2021training,hendrycks2021measuring,austin2021program,lu2023dynamic}.
However, these math reasoning benchmarks typically do not require specialized domain knowledge, a notable shortcoming when considering the practical applications of LLMs. Therefore, recent work has investigated the LLMs' capabilities in knowledge-intensive problem solving. For example, \citet{chen-etal-2023-theoremqa} collected a theorem-driven question-answering dataset, designed to evaluate AI models' ability to apply theorems in solving challenging science problems. MMMU~\cite{yue2023mmmu} and MathVista~\cite{lu2023mathvista} include examples that require complex multimodal reasoning in expert domains. 
Different from this recent work, which focuses on benchmarking LLM performance, our work also constructs a finance-domain knowledge bank, investigating various knowledge integration strategies to enhance knowledge-intensive problem solving. 
Moreover, \ours also requires LLMs to understand and interpret tabular data in expert domains to solve the problems. 

\section{Conclusion}
This paper introduces \ours, a benchmark aimed at assessing LLMs in knowledge-intensive math reasoning. Our comprehensive evaluations of \nmodel LLMs, using both CoT and PoT prompting methods, identify significant areas where LLMs need to enhance their specialized knowledge for complex problem-solving in expert domains. Additionally, our knowledge augmentation analysis indicates that integrating domain-specific knowledge can improve LLMs' problem-solving abilities. We believe this research provides valuable insights into advancing LLMs within expert domains.

\section*{Limitations}
In this work, we propose \ours and conduct comprehensive analysis of different LLMs' capabilities in solving knowledge-intensive math reasoning problems in finance domains. 
However, there are still some limitations: (1) Our method for extracting final answer from model output is still not perfect. In some cases, this methods fails to locate the answer, leading to the reported accuracy being an approximate lower bound.
Moreover, as the extracted answer can be in a different format than the ground truth, we apply rule-based methods to measure the exact match between the two values, which could introduce around 2\% errors based on our case studies.
(2) In our experiment, we regard tables in the question as textual input. However, in real-world scenarios, tabular data might appear as images, where people cannot obtain its textual content directly. In these cases, OCR tools to extract table content~\cite{du2020pp} or LLMs with vision capabilities~\cite{2023GPT4VisionSC, yue2023mmmu, lu2023mathvista} may be required.
(3) 
% Among recently released finance-specific LLMs~\cite{Wu2023BloombergGPTAL, yang2023fingpt, xie2023pixiu}, we only evaluate FinMA, as it is the only work with a checkpoint available at HuggingFace and compatible with the vllm framework. 
% However, FinMA uses Llama-1 as its base models, performing worse than more current LLMs (\autoref{tab:results}). 
Due to computational resource constraints, we do not tune LLMs on a large-scale finance-domain data ourselves~\cite{xie2023pixiu, xie2024finben}. However, we believe that training on finance data can help improve knowledge-intensive problem solving in finance domains.
\section*{Acknowledgement}
We are grateful for the compute support provided by Microsoft Research’s Accelerate Foundation Models Research (AFMR) program.
We would also like to thank the anonymous reviewers and area chairs for constructive discussions and feedback. Hongjun Liu and Chen Zhao are supported by Shanghai Frontiers Science Center of Artificial Intelligence and Deep Learning, NYU Shanghai.
\bibliography{anthology, custom, llms}

\appendix

\clearpage
\section{Appendix}\label{appendix:dataset}
\begin{table}[h]
\centering
\small
\begin{tabular}{lccc}
\toprule
\textbf{Annotation Quality}    & \textbf{\%S $\geq$ 4}\\
\midrule
Question Fluency  & 98.0 \\
Question Correctness &  95.3 \\
\noalign{\vskip 0.5ex}\hdashline\noalign{\vskip 0.5ex}
Knowledge Relevance  &  94.1 \\
Textual Definition Fluency & 93.0\\
Textual Definition Correctness & 94.7\\
Math Formula Correctness & 88.0\\
\noalign{\vskip 0.5ex}\hdashline\noalign{\vskip 0.5ex}
Final Answer Correctness & 98.0\\
Python Solution Correctness & 96.0\\
Variable Name Meaningfulness & 87.7\\
Comment Comprehensiveness & 83.8\\
\bottomrule
\end{tabular}

\caption{Human evaluation over 200 samples of \ours. Three internal evaluators were asked to rate the samples on a scale of 1 to 5 individually. We report percent of samples that have an average score $\geq$ 4 to indicate the annotation quality of \ours.}
\label{tab:annotation_aggrement}
\end{table}

\begin{figure}[h]
\begin{tcolorbox}[colback=black!7.5!white, colframe=black!80!white, title=Program-of-Thought Prompt Used, fontupper=\footnotesize, fonttitle=\footnotesize]
\texttt{[System Input]}: \vspace{2pt}\\
You are a financial expert, you are supposed to generate a Python program to answer the given question. The returned value of the program is supposed to be the answer. Here is an example of the Python program:
\begin{verbatim}
```python
def solution():
    # Define variables name and value
    ...
    
    # Do math calculation
    ...
    
    # return answer
    return answer
```
\end{verbatim}
\texttt{[User Input]}: \vspace{2pt}\\
Question: \{question\}\\
\newline
Generate a Python program to answer the given question.
Continue your output:
\begin{verbatim}
```python
def solution():
    # Define variables name and value
\end{verbatim}
\end{tcolorbox}

\caption{Example of \emph{zero}-shot PoT prompt used.}
\label{fig:pot}
\end{figure}
\begin{figure}[!t]
\begin{tcolorbox}[colback=black!7.5!white, colframe=black!80!white, title=Chain-of-Thought Prompt, fontupper=\footnotesize, fonttitle=\footnotesize]
\texttt{[System Input]}: \vspace{2pt}\\
You are a financial expert, you are supposed to answer the given question. You need to first think through the problem step by step, documenting each necessary step. Then you are required to conclude your response with the final answer in your last sentence as ``Therefore, the answer is \{final answer\}''. The final answer should be a numeric value. \\
\newline
\texttt{[User Input]}: \vspace{2pt}\\
Question: \{question\}\\
\newline
Relevant Knowledge:\\
\{knowledge\}\\
\newline
Let's think step by step to answer the given question.
\end{tcolorbox}

\caption{Example of \emph{zero}-shot CoT prompt used for knowledge augmentation analysis in Section~\ref{sec:knowledge-analysis}. }
\label{fig:knowledge_aug_cot}
\end{figure}
\begin{table}[!t]
\resizebox{\linewidth}{!}{%
\addtolength{\tabcolsep}{-0.1em}
\small
\centering
\begin{tabular}{p{9cm}}
\toprule
\texttt{[validation-69]}\\

What is the gross domestic product for country B in 2015, according to the provided data?\newline
\begin{tabular}[t]{lr}
    \toprule
    Account Item & (\$ trillions) \\ 
    \midrule
    Consumption & 20.0 \\
    Capital consumption allowance & 1.9 \\
    Government spending	 & 4.2\\
    Imports	& 2.2 \\
    Gross private domestic investment & 6.0\\
    Exports	& 1.8 \\
    \bottomrule
\end{tabular} 
\\
\vspace{10pt}

\textbf{Model Output} \\
\texttt{(..abbreviate..)}

Net Investment = Gross private domestic investment \textcolor{red}{- Capital consumption allowance} = 6.0 - 1.9 = 4.1 trillion \newline

Net Exports = Exports - Imports = 1.8 - 2.2 = -0.4 trillion (note: negative value indicates a trade deficit) \newline

We can plug in the values into the GDP formula:
GDP = Consumption + Net Investment + Government Spending + Net Exports = 20.0 + 4.1 + 4.2 - 0.4 = \textcolor{red}{28.9} trillion
\\
\vspace{10pt}
\textbf{Error Explanation} \\
Misinterpretation of the GDP components: The model incorrectly subtracts the capital consumption allowance to derive net investment, but GDP should use gross investment. \newline
Additionally, the math computation in the last step is incorrect \\
\bottomrule
\end{tabular}
}
\caption{Example of knowledge misinterpretation error made by Llama-3-70B.}
\label{tab:error_analysis}
\end{table}

\begin{table*}[!t]
\centering
\resizebox{0.95\textwidth}{!}{%
\renewcommand{\arraystretch}{1.1}
\begin{tabular}{llllp{10cm}}
\toprule
Model & Organization & Size & Notes & Source \\
\midrule
GPT-4-Turbo & OpenAI & -- & & \texttt{gpt-4o-2024-05-13} \\
GPT-4o & OpenAI & -- & & \texttt{gpt-4-turbo-2024-04-09} \\
GPT-3.5-Turbo & OpenAI & -- & & \texttt{gpt-3.5-turbo-0125} \\
\noalign{\vskip 0.5ex}\hdashline\noalign{\vskip 0.5ex}

Claude-3.5-Sonnet & Anthropic & -- & & \texttt{claude-3-5-sonnet-20240620} \\
Claude-3-Opus & Anthropic & -- & & \texttt{claude-3-opus-20240229} \\
Claude-3-Sonnet & Anthropic & -- & & \texttt{claude-3-sonnet-20240229} \\
Claude-3-Haiku & Anthropic & -- & & \texttt{claude-3-haiku-20240307} \\
\noalign{\vskip 0.5ex}\hdashline\noalign{\vskip 0.5ex}

Gemini-1.5-Pro & Google & -- & & \texttt{gemini-1.5-pro} \\
Gemini-1.5-Flash & Google & -- & & \texttt{gemini-1.5-flash} \\

\midrule \midrule
Qwen2 & Alibaba & 7 \& 72B & & \texttt{Qwen/Qwen2-*B-Instruct} \\

\noalign{\vskip 0.5ex}\hdashline\noalign{\vskip 0.5ex}
Llama-2 & Meta & 7 \& 70B & & \texttt{meta-llama/Llama-2-*b-chat-hf} \\
Llama-3 & Meta & 8 \& 70B & & \texttt{meta-llama/Meta-Llama-3-*B-Instruct} \\
Llama-3.1 & Meta & 8 \& 70B \& 405B & & \texttt{meta-llama/Meta-Llama-3.1-*B-Instruct} \\

\noalign{\vskip 0.5ex}\hdashline\noalign{\vskip 0.5ex}
Gemma-1 & Google & 2 \& 7B & & \texttt{google/gemma-b-it} \\
Gemma-2 & Google & 9B & & \texttt{google/gemma-2-9b-it} \\

\noalign{\vskip 0.5ex}\hdashline\noalign{\vskip 0.5ex}
Mistral-v0.3 & Mixtral AI & 7B & & \texttt{mistralai/Mistral-7B-Instruct-v0.3} \\
Mistral-Nemo & Mixtral AI & 12B & & \texttt{mistralai/Mistral-Nemo-Instruct-2407} \\
Mistral-Large & Mixtral AI & 123B & & \texttt{mistralai/Mistral-Large-Instruct-2407} \\
Mathstral & Mixtral AI & 7B & Math-Specific & \texttt{mistralai/Mathstral-7B-v0.1} \\
Mixtral & Mixtral AI & 46 \& 141B & MoE & \texttt{mistralai/Mixtral--Instruct-v0.1} \\
Codestral & Mixtral AI & 22B & Code-Specific & \texttt{mistralai/Codestral-22B-v0.1} \\

\noalign{\vskip 0.5ex}\hdashline\noalign{\vskip 0.5ex}
DeepSeek-Math & DeepSeek & 7B & Math-Specific & \texttt{deepseek-ai/deepseek-math-7b-instruct} \\
DeepSeek-Coder-V1 & DeepSeek & 33B & Code-Specific & \texttt{deepseek-ai/deepseek-coder-33b-instruct} \\
\\

DeepSeek-V2 & DeepSeek & 16 \& 236B & MoE & \texttt{deepseek-ai/DeepSeek-V2-Lite-Chat}. We use the official API provided by DeepSeek for \texttt{deepseek-ai/DeepSeek-V2-Chat} \\
\\
DeepSeek-Coder-V2 & DeepSeek & 16 \& 236B & MoE & \texttt{deepseek-ai/DeepSeek-Coder-V2-Lite-Instruct}. We use the official API provided by DeepSeek for \texttt{deepseek-ai/DeepSeek-Coder-V2-Instruct}\\

\noalign{\vskip 0.5ex}\hdashline\noalign{\vskip 0.5ex}
Yi-1.5 & 01 AI & 9 \& 34B & & \texttt{01-ai/Yi-1.5-34B-Chat} \\
\noalign{\vskip 0.5ex}\hdashline\noalign{\vskip 0.5ex}
Phi-3-Medium & Microsoft & 14B & & \texttt{microsoft/Phi-3-medium-4k-instruct} \\
Phi-3-Mini & Microsoft & 3B & & \texttt{microsoft/Phi-3-mini-4k-instruct} \\

\noalign{\vskip 0.5ex}\hdashline\noalign{\vskip 0.5ex}
GLM-4 & THUDM & 9B & & \texttt{THUDM/glm-4-9b-chat} \\

\noalign{\vskip 0.5ex}\hdashline\noalign{\vskip 0.5ex}
DBRX & Databricks & 132B & MoE & \texttt{databricks/dbrx-instruct} \\

\noalign{\vskip 0.5ex}\hdashline\noalign{\vskip 0.5ex}
C4AI Command R+ & Cohere & 104B & & \texttt{CohereForAI/c4ai-command-r-plus} \\

\noalign{\vskip 0.5ex}\hdashline\noalign{\vskip 0.5ex}
InternLM2 & InternLM & 7B & & \texttt{internlm/internlm2-chat-7b} \\
InternLM2-Math-Plus & InternLM & 7B & Math-Specific & \texttt{internlm/internlm2-math-plus-7b} \\

\noalign{\vskip 0.5ex}\hdashline\noalign{\vskip 0.5ex}
WizardLM-2 & WizardLM Team & 7B & & \texttt{lucyknada/microsoft\_WizardLM-2-7B} \\
WizardMath & WizardLM Team & 7B & Math-Specific & \texttt{WizardLMTeam/WizardMath-7B-V1.1} \\
WizardCoder & WizardLM Team & 33B & Code-Specific & \texttt{WizardLMTeam/WizardCoder-33B-V1.1} \\
WizardLM-2 (MoE) & WizardLM Team & 141B & MoE & \texttt{alpindale/WizardLM-2-8x22B} \\

\noalign{\vskip 0.5ex}\hdashline\noalign{\vskip 0.5ex}
Aya-23 & Cohere & 8 \& 35B & & \texttt{CohereForAI/aya-23-*B} \\

\noalign{\vskip 0.5ex}\hdashline\noalign{\vskip 0.5ex}
StarCoder2 & BigCode & 15B & Code-Specific & \texttt{bigcode/starcoder2-15b-instruct-v0.1} \\
\bottomrule
\end{tabular}
}
\caption{Details of the organization and model source (\ie model version for proprietary models, and Huggingface model name for open-source models) for the LLMs evaluated in \ours. }
\label{tab:model-detail}
\end{table*}

\begin{table*}[!t]
\centering
\resizebox{\textwidth}{!}{%
\renewcommand{\arraystretch}{1.1}
\addtolength{\tabcolsep}{-0.2em}
\begin{tabular}{lrl:rrrrrrrrrrrrrrrrrrrrr:rr}
\toprule
\multirow{2}{*}{\textbf{Model}} & \multirow{2}{*}{\textbf{Size}}  & \multirow{2}{*}{\textbf{Notes}} & \multicolumn{2}{c}{\textbf{Quantitative}} && \multicolumn{2}{c}{\textbf{Derivatives}} && \multicolumn{2}{c}{\textbf{Accounting}} && \multicolumn{2}{c}{\textbf{Management}} && \multicolumn{2}{c}{\textbf{Portfolio}} && \multicolumn{2}{c}{\textbf{Economics}} && \multicolumn{2}{c}{\textbf{Corporate}} && \multicolumn{2}{c}{\textbf{Avg.}} \\
\cmidrule(lr){4-5} \cmidrule(lr){7-8} \cmidrule(lr){10-11} \cmidrule(lr){13-14} \cmidrule(lr){16-17} \cmidrule(lr){19-20} \cmidrule(lr){22-23} \cmidrule(lr){25-26}
& & & PoT & CoT && PoT & CoT && PoT & CoT && PoT & CoT &&  PoT & CoT  &&PoT & CoT &&PoT & CoT && PoT & \textbf{CoT}\\
\midrule
Close-book \\
\quad Expert &  & &  & & & &   &  &  &  &  &  &  &  &  &  &  &  &  &  &  &  &  & \multicolumn{2}{c}{73.0} \\
\quad Non-Expert &  & & & &  &  &  &  &  &  &  &  &  &  &  &  &  &  &  &  &  &  &  & \multicolumn{2}{c}{58.0} \\
\noalign{\vskip 0.5ex}\hdashline\noalign{\vskip 0.5ex}
Open-book \\
\quad Expert &  & &  & & & &   &  &  &  &  &  &  &  &  &  &  &  &  &  &  &  &  & \multicolumn{2}{c}{92.0} \\
\quad Non-Expert &  & & & & &   &  &  &  &  &  &  &  &  &  &  &  &  &  &  &  &  &  & \multicolumn{2}{c}{84.0} \\

\midrule
\multicolumn{26}{c}{\emph{\textbf{Proprietary LLMs}}} \\\noalign{\vskip 0.5ex}

GPT-4o &  &  &\cellcolor{blue!35}75.0 & 45.8 &  & \cellcolor{blue!35}58.8 & \cellcolor{blue!35}55.4 &  & \cellcolor{blue!5}60.3 & \cellcolor{blue!20}69.4 &  & \cellcolor{blue!35}82.9 & \cellcolor{blue!35}66.3 &  & \cellcolor{blue!35}77.8 & \cellcolor{blue!5}69.4 &  & \cellcolor{blue!20}62.5 & \cellcolor{blue!35}58.9 &  & \cellcolor{blue!35}67.0 & \cellcolor{blue!20}56.9 &  & \cellcolor{blue!35}\underline{67.0} & \cellcolor{blue!35}60.9 \\
Claude-3.5-Sonnet &  &  &\cellcolor{blue!20}73.6 & \cellcolor{blue!20}55.6 &  & \cellcolor{blue!20}54.1 & \cellcolor{blue!20}51.8 &  & \cellcolor{blue!35}66.7 & \cellcolor{blue!35}69.9 &  & \cellcolor{blue!5}75.6 & \cellcolor{blue!5}63.9 &  & \cellcolor{blue!5}72.2 & \cellcolor{blue!20}72.2 &  & \cellcolor{blue!35}64.3 & \cellcolor{blue!20}58.9 &  & \cellcolor{blue!20}62.4 & \cellcolor{blue!35}60.6 &  & \cellcolor{blue!20}\underline{64.8} & \cellcolor{blue!20}60.6 \\
Claude-3-Opus &  &  &66.7 & \cellcolor{blue!35}56.9 &  & \cellcolor{blue!5}53.5 & \cellcolor{blue!5}45.2 &  & \cellcolor{blue!20}62.1 & 59.8 &  & \cellcolor{blue!20}79.5 & \cellcolor{blue!20}64.9 &  & \cellcolor{blue!20}72.2 & \cellcolor{blue!35}83.3 &  & 51.8 & 46.4 &  & \cellcolor{blue!5}59.6 & 45.0 &  & \cellcolor{blue!5}\underline{62.9} & \cellcolor{blue!5}54.7 \\
GPT-4-Turbo &  &  &59.7 & 38.9 &  & 49.8 & 42.2 &  & 50.7 & \cellcolor{blue!5}64.8 &  & 72.2 & 56.6 &  & 61.1 & 50.0 &  & 57.1 & 44.6 &  & 50.5 & \cellcolor{blue!5}47.7 &  & \underline{56.2} & 50.9 \\
Gemini-1.5-Pro &  &  &68.1 & \cellcolor{blue!5}50.0 &  & 53.1 & 30.7 &  & 56.6 & 55.2 &  & 69.8 & 57.6 &  & 58.3 & 63.9 &  & 51.8 & \cellcolor{blue!5}55.4 &  & 50.5 & 44.0 &  & \underline{58.2} & 47.0 \\
GPT-4o-Mini &  &  &65.3 & 36.1 &  & 46.9 & 29.7 &  & 48.4 & 47.5 &  & 69.3 & 46.8 &  & 50.0 & 38.9 &  & \cellcolor{blue!5}57.1 & 41.1 &  & 51.4 & 45.9 &  & \underline{54.3} & 40.3 \\
Gemini-1.5-Flash &  &  &\cellcolor{blue!5}69.4 & 33.3 &  & 43.6 & 28.7 &  & 52.0 & 48.9 &  & 67.8 & 49.8 &  & 58.3 & 61.1 &  & 50.0 & 37.5 &  & 47.7 & 34.9 &  & \underline{53.6} & 40.1 \\
Claude-3-Sonnet &  &  &59.7 & 37.5 &  & 37.0 & 28.4 &  & 48.0 & 43.4 &  & 66.8 & 48.8 &  & 47.2 & 55.6 &  & 48.2 & 33.9 &  & 48.6 & 35.8 &  & \underline{49.4} & 38.6 \\
Claude-3-Haiku &  &  &34.7 & 31.9 &  & 19.8 & 26.4 &  & 33.8 & 43.4 &  & 44.9 & 41.5 &  & 41.7 & 44.4 &  & 25.0 & 33.9 &  & 36.7 & 30.3 &  & 32.0 & 35.1 \\
GPT-3.5-Turbo &  &  &47.2 & 25.0 &  & 24.4 & 16.5 &  & 29.2 & 29.2 &  & 51.2 & 33.2 &  & 27.8 & 22.2 &  & 37.5 & 21.4 &  & 32.1 & 23.8 &  & \underline{34.3} & 24.6 \\

\midrule
\multicolumn{26}{c}{\emph{\textbf{Open-source LLMs}}} \\\noalign{\vskip 0.5ex}

DeepSeek-Coder-V2 & 236B & Code &\cellcolor{blue!20}38.5 & 41.0 &  & \cellcolor{blue!20}62.5 & \cellcolor{blue!20}66.7 &  & \cellcolor{blue!20}47.6 & \cellcolor{blue!35}46.0 &  & \cellcolor{blue!5}75.0 & \cellcolor{blue!35}61.1 &  & 33.3 & \cellcolor{blue!35}44.4 &  & \cellcolor{blue!35}79.0 & \cellcolor{blue!35}52.6 &  & \cellcolor{blue!35}60.0 & \cellcolor{blue!35}50.0 &  & \cellcolor{blue!35}\underline{55.5} & \cellcolor{blue!35}51.0 \\
DeepSeek-V2 & 236B & MoE &\cellcolor{blue!5}38.5 & \cellcolor{blue!5}46.2 &  & \cellcolor{blue!35}66.7 & \cellcolor{blue!35}79.2 &  & \cellcolor{blue!35}47.6 & \cellcolor{blue!20}44.4 &  & 75.0 & \cellcolor{blue!20}52.8 &  & \cellcolor{blue!5}44.4 & \cellcolor{blue!5}33.3 &  & \cellcolor{blue!5}73.7 & \cellcolor{blue!5}47.4 &  & 30.0 & \cellcolor{blue!5}40.0 &  & \cellcolor{blue!20}\underline{54.5} & \cellcolor{blue!20}50.0 \\
Llama-3.1 & 405B &  &\cellcolor{blue!35}41.0 & \cellcolor{blue!35}51.3 &  & \cellcolor{blue!5}54.2 & \cellcolor{blue!5}54.2 &  & \cellcolor{blue!5}46.0 & \cellcolor{blue!5}33.3 &  & \cellcolor{blue!35}80.6 & 47.2 &  & 33.3 & 11.1 &  & 68.4 & 31.6 &  & \cellcolor{blue!5}40.0 & \cellcolor{blue!20}50.0 &  & \cellcolor{blue!5}\underline{53.5} & \cellcolor{blue!5}41.5 \\
Mistral-Large & 123B &  &38.5 & \cellcolor{blue!20}51.3 &  & 50.0 & 45.8 &  & 38.1 & 30.2 &  & \cellcolor{blue!20}77.8 & 36.1 &  & \cellcolor{blue!20}44.4 & 22.2 &  & \cellcolor{blue!20}73.7 & 31.6 &  & 40.0 & 10.0 &  & \underline{50.5} & 36.0 \\
Llama-3.1 & 70B &  &25.6 & 38.5 &  & 41.7 & 50.0 &  & 33.3 & 19.0 &  & 66.7 & \cellcolor{blue!5}47.2 &  & \cellcolor{blue!35}55.6 & 22.2 &  & 68.4 & \cellcolor{blue!20}47.4 &  & 30.0 & 30.0 &  & \underline{43.0} & 35.0 \\
Qwen2 & 72B &  &23.1 & 33.3 &  & 29.2 & 45.8 &  & 23.8 & 22.2 &  & 41.7 & 47.2 &  & 22.2 & 22.2 &  & 68.4 & 31.6 &  & 10.0 & 30.0 &  & 31.0 & 33.0 \\
Llama-3 & 70B &  &33.3 & 38.5 &  & 37.5 & 45.8 &  & 33.3 & 22.2 &  & 66.7 & 33.3 &  & 33.3 & 22.2 &  & 68.4 & 31.6 &  & 30.0 & 30.0 &  & \underline{43.0} & 31.5 \\
Phi-3-Medium & 14B &  &20.5 & 28.2 &  & 37.5 & 50.0 &  & 25.4 & 19.0 &  & 55.6 & 41.7 &  & 22.2 & 22.2 &  & 52.6 & 42.1 &  & \cellcolor{blue!20}40.0 & 20.0 &  & \underline{34.5} & 31.0 \\
DeepSeek-Coder-V2-Lite & 16B & Code &23.1 & 28.2 &  & 25.0 & 29.2 &  & 27.0 & 23.8 &  & 50.0 & 33.3 &  & 11.1 & 22.2 &  & 42.1 & 31.6 &  & 10.0 & 20.0 &  & \underline{30.0} & 27.5 \\
Mixtral-8x22B & 141B & MoE &7.7 & 28.2 &  & 4.2 & 45.8 &  & 0.0 & 19.0 &  & 25.0 & 30.6 &  & 11.1 & 11.1 &  & 21.0 & 26.3 &  & 10.0 & 20.0 &  & 9.5 & 26.5 \\
Yi-1.5 & 9B &  &10.3 & 30.8 &  & 20.8 & 25.0 &  & 11.1 & 14.3 &  & 36.1 & 30.6 &  & 33.3 & 22.2 &  & 26.3 & 26.3 &  & 10.0 & 20.0 &  & 19.0 & 23.5 \\
Yi-1.5 & 34B &  &10.3 & 25.6 &  & 12.5 & 25.0 &  & 14.3 & 9.5 &  & 30.6 & 33.3 &  & 0.0 & 22.2 &  & 26.3 & 21.0 &  & 10.0 & 10.0 &  & 16.5 & 20.5 \\
WizardLM-2 & 141B & MoE &15.4 & 28.2 &  & 41.7 & 29.2 &  & 17.5 & 11.1 &  & 52.8 & 19.4 &  & 0.0 & 11.1 &  & 47.4 & 15.8 &  & 10.0 & 40.0 &  & \underline{28.0} & 20.0 \\
Gemma-2 & 9B &  &23.1 & 20.5 &  & 25.0 & 29.2 &  & 20.6 & 11.1 &  & 36.1 & 33.3 &  & 11.1 & 0.0 &  & 42.1 & 26.3 &  & 10.0 & 10.0 &  & \underline{25.5} & 20.0 \\
GLM-4 & 9B &  &18.0 & 18.0 &  & 25.0 & 37.5 &  & 14.3 & 17.5 &  & 27.8 & 13.9 &  & 22.2 & 0.0 &  & 26.3 & 21.0 &  & 10.0 & 30.0 &  & \underline{20.0} & 19.5 \\
DBRX & 132B & MoE &12.8 & 28.2 &  & 16.7 & 20.8 &  & 4.8 & 9.5 &  & 13.9 & 16.7 &  & 11.1 & \cellcolor{blue!20}33.3 &  & 21.0 & 21.0 &  & 0.0 & 20.0 &  & 11.0 & 18.5 \\
Phi-3-Mini & 3B &  &18.0 & 23.1 &  & 25.0 & 25.0 &  & 11.1 & 7.9 &  & 30.6 & 16.7 &  & 0.0 & 22.2 &  & 31.6 & 15.8 &  & 20.0 & 20.0 &  & \underline{19.5} & 16.5 \\
DeepSeek-Math & 7B & Math &0.0 & 18.0 &  & 0.0 & 20.8 &  & 0.0 & 12.7 &  & 2.8 & 13.9 &  & 0.0 & 11.1 &  & 0.0 & 31.6 &  & 0.0 & 10.0 &  & 0.5 & 16.5 \\
Mathstral & 7B & Math &18.0 & 23.1 &  & 20.8 & 25.0 &  & 11.1 & 11.1 &  & 27.8 & 19.4 &  & 11.1 & 11.1 &  & 36.8 & 15.8 &  & 0.0 & 0.0 &  & \underline{18.5} & 16.5 \\
Llama-3.1 & 8B &  &23.1 & 12.8 &  & 20.8 & 33.3 &  & 7.9 & 6.4 &  & 38.9 & 27.8 &  & 11.1 & 11.1 &  & 31.6 & 15.8 &  & 20.0 & 10.0 &  & \underline{21.0} & 16.0 \\
Qwen2 & 7B &  &7.7 & 28.2 &  & 8.3 & 12.5 &  & 6.4 & 12.7 &  & 8.3 & 16.7 &  & 0.0 & 11.1 &  & 10.5 & 15.8 &  & 0.0 & 0.0 &  & 7.0 & 16.0 \\
Mistral-Nemo & 12B &  &5.1 & 20.5 &  & 12.5 & 25.0 &  & 4.8 & 9.5 &  & 41.7 & 16.7 &  & 11.1 & 11.1 &  & 47.4 & 21.0 &  & 10.0 & 0.0 &  & \underline{17.0} & 15.5 \\
C4AI Command R+ & 104B &  &5.1 & 18.0 &  & 4.2 & 25.0 &  & 0.0 & 7.9 &  & 5.6 & 11.1 &  & 0.0 & 0.0 &  & 5.3 & 15.8 &  & 10.0 & 0.0 &  & 3.5 & 12.5 \\
Codestral & 22B & Code &18.0 & 7.7 &  & 25.0 & 20.8 &  & 17.5 & 9.5 &  & 50.0 & 16.7 &  & 22.2 & 0.0 &  & 42.1 & 15.8 &  & 10.0 & 0.0 &  & \underline{26.5} & 11.5 \\
Mixtral-8x7B-v0.1 & 46B & MoE &0.0 & 10.3 &  & 0.0 & 16.7 &  & 0.0 & 6.4 &  & 0.0 & 22.2 &  & 0.0 & 0.0 &  & 0.0 & 15.8 &  & 0.0 & 0.0 &  & 0.0 & 11.5 \\
Llama-3 & 8B &  &18.0 & 10.3 &  & 20.8 & 0.0 &  & 11.1 & 7.9 &  & 25.0 & 13.9 &  & 11.1 & 11.1 &  & 26.3 & 21.0 &  & 0.0 & 0.0 &  & \underline{17.0} & 9.5 \\
WizardMath & 7B & Math &7.7 & 10.3 &  & 12.5 & 16.7 &  & 4.8 & 7.9 &  & 13.9 & 5.6 &  & 0.0 & 11.1 &  & 15.8 & 10.5 &  & 0.0 & 10.0 &  & 8.5 & 9.5 \\
InternLM2-Math-Plus & 7B & Math &12.8 & 7.7 &  & 8.3 & 16.7 &  & 6.4 & 7.9 &  & 13.9 & 8.3 &  & 11.1 & 0.0 &  & 21.0 & 10.5 &  & 0.0 & 20.0 &  & \underline{10.5} & 9.5 \\
DeepSeek-V2-Lite & 16B & MoE &7.7 & 10.3 &  & 4.2 & 8.3 &  & 1.6 & 9.5 &  & 2.8 & 5.6 &  & 0.0 & 0.0 &  & 10.5 & 10.5 &  & 0.0 & 10.0 &  & 4.0 & 8.5 \\
WizardLM-2 & 7B &  &12.8 & 5.1 &  & 16.7 & 12.5 &  & 6.4 & 6.4 &  & 25.0 & 11.1 &  & 0.0 & 0.0 &  & 21.0 & 15.8 &  & 0.0 & 0.0 &  & \underline{13.0} & 8.0 \\
Llama-2 & 70B &  &15.4 & 7.7 &  & 16.7 & 8.3 &  & 4.8 & 4.8 &  & 11.1 & 5.6 &  & 0.0 & 0.0 &  & 15.8 & 21.0 &  & 10.0 & 0.0 &  & \underline{10.5} & 7.0 \\
Aya-23 & 35B &  &0.0 & 5.1 &  & 0.0 & 12.5 &  & 0.0 & 6.4 &  & 0.0 & 5.6 &  & 0.0 & 0.0 &  & 0.0 & 15.8 &  & 0.0 & 0.0 &  & 0.0 & 7.0 \\
Mistral-v0.3 & 7B &  &0.0 & 12.8 &  & 0.0 & 8.3 &  & 4.8 & 3.2 &  & 2.8 & 5.6 &  & 0.0 & 0.0 &  & 5.3 & 10.5 &  & 0.0 & 0.0 &  & 2.5 & 6.5 \\
StarCoder2 & 15B & Code &5.1 & 10.3 &  & 8.3 & 8.3 &  & 7.9 & 7.9 &  & 36.1 & 2.8 &  & 33.3 & 0.0 &  & 52.6 & 0.0 &  & 0.0 & 10.0 &  & \underline{17.5} & 6.5 \\
InternLM2 & 7B &  &7.7 & 7.7 &  & 16.7 & 4.2 &  & 3.2 & 7.9 &  & 11.1 & 2.8 &  & 0.0 & 0.0 &  & 26.3 & 5.3 &  & 0.0 & 0.0 &  & \underline{9.0} & 5.5 \\
DeepSeek-Coder-V1 & 33B & Code &2.6 & 10.3 &  & 8.3 & 4.2 &  & 3.2 & 3.2 &  & 13.9 & 8.3 &  & 0.0 & 0.0 &  & 21.0 & 0.0 &  & 0.0 & 0.0 &  & \underline{7.0} & 5.0 \\
WizardCoder & 33B & Code &18.0 & 2.6 &  & 20.8 & 8.3 &  & 6.4 & 4.8 &  & 27.8 & 5.6 &  & 0.0 & 11.1 &  & 21.0 & 0.0 &  & 10.0 & 0.0 &  & \underline{15.5} & 4.5 \\
Aya-23 & 8B &  &0.0 & 10.3 &  & 0.0 & 8.3 &  & 0.0 & 3.2 &  & 0.0 & 0.0 &  & 0.0 & 0.0 &  & 0.0 & 5.3 &  & 0.0 & 0.0 &  & 0.0 & 4.5 \\
Llama-2 & 7B &  &2.6 & 2.6 &  & 4.2 & 0.0 &  & 1.6 & 6.4 &  & 2.8 & 0.0 &  & 0.0 & 0.0 &  & 5.3 & 0.0 &  & 0.0 & 10.0 &  & 2.5 & 3.0 \\
Gemma-1 & 2B &  &5.1 & 2.6 &  & 4.2 & 4.2 &  & 1.6 & 6.4 &  & 0.0 & 0.0 &  & 0.0 & 0.0 &  & 0.0 & 0.0 &  & 0.0 & 0.0 &  & 2.0 & 3.0 \\
Gemma-1 & 7B &  &5.1 & 5.1 &  & 8.3 & 4.2 &  & 1.6 & 4.8 &  & 0.0 & 0.0 &  & 0.0 & 0.0 &  & 0.0 & 0.0 &  & 0.0 & 0.0 &  & 2.5 & 3.0 \\

\bottomrule
\end{tabular}
}
\caption{Results of Chain-of-Thought and Program-of-Thought prompting on the \dev set of \ours.
% We report the accuracy over different fine-grained question types. 
We select the most recent version as of July 5, 2024, for each model.
We use average Accuracy using CoT prompting as the ranking indicator of model performance. \underline{Numbers} underscored indicate that models with PoT prompting achieves better results than with CoT prompting. }
% Models marked with $^*$ are found challenging to instruct for generating \emph{Python-format solutions} using PoT prompting; therefore, we only report their performance with CoT prompting approach.}
\label{tab:results-dev}
\end{table*}
\begin{figure*}[!t]
    \centering
    \includegraphics[width = \linewidth]{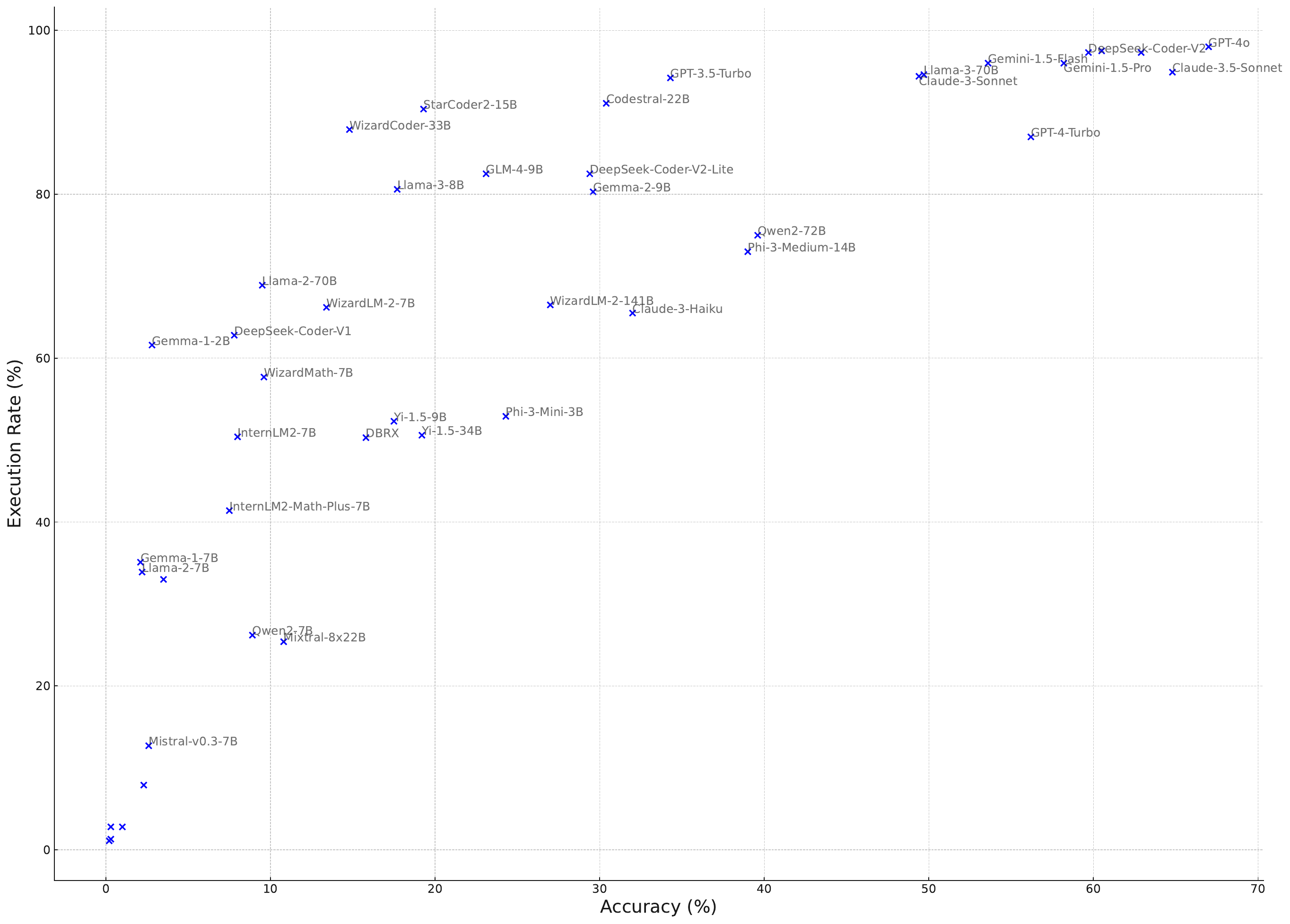}
    \caption{Relationship between execution rate and accuracy across different LLMs with PoT prompting on test set.}
    \label{fig:execution-rate}
\end{figure*}

\end{document}